\documentclass{article}
     \PassOptionsToPackage{numbers, compress}{natbib}

     \usepackage[preprint]{neurips_2022}

\usepackage[utf8]{inputenc} 
\usepackage[T1]{fontenc}    
\usepackage{hyperref}       
\usepackage{url}            
\usepackage{booktabs}       
\usepackage{amsfonts}       
\usepackage{nicefrac}       
\usepackage{microtype}      
\usepackage{xcolor}         
\usepackage{amsmath}
\usepackage{graphicx}
\usepackage{subfig}
\usepackage{multirow}
\allowdisplaybreaks[4]

\title{Reinforced Swin-Convs Transformer for Underwater Image Enhancement}

\author{%
	Tingdi Ren \\
	Ningbo University\\
	\texttt{tingdiren@gmail.com} \\
	\And
	Haiyong Xu\thanks{This work was supported in part by the Natural Science Foundation of China (62171243, 61871247,  62071266,  61931022, 61671412), in part by the Zhejiang Natural Science Foundation of China (LY21F010003,  LQ20F010002,  LY21F010014). Corresponding author: Haiyong Xu } \\
	Ningbo University\\
	\texttt{xuhaiyong@nbu.edu.cn} \\
	\And
	Gangyi Jiang \\
	Ningbo University\\
	\texttt{jianggangyi@nbu.edu.cn} \\  
	\And  
	Mei Yu \\
	Ningbo University\\
	\texttt{yumei@nbu.edu.cn} \\ 
	\And  
	Ting Luo \\
	Ningbo University\\
	\texttt{luoting@nbu.edu.cn} \\ 
}

\begin{document}

\maketitle

\begin{abstract}

Underwater Image Enhancement (UIE) technology aims to tackle the challenge of restoring the degraded underwater images due to light absorption and scattering. To address problems, a novel U-Net based Reinforced Swin-Convs Transformer for the Underwater Image Enhancement method (URSCT-UIE) is proposed. Specifically, with the deficiency of U-Net based on pure convolutions, we embedded the Swin Transformer into U-Net for improving the ability to capture the global dependency. Then, given the inadequacy of the Swin Transformer capturing the local attention, the reintroduction of convolutions may capture more local attention. Thus, we provide an ingenious manner for the fusion of convolutions and the core attention mechanism to build a Reinforced Swin-Convs Transformer Block (RSCTB) for capturing more local attention, which is reinforced in the channel and the spatial attention of the Swin Transformer. Finally, the experimental results on available datasets demonstrate that the proposed URSCT-UIE achieves state-of-the-art performance compared with other methods in terms of both subjective and objective evaluations.  The code will be released on GitHub after acceptance.
\end{abstract}

\section{Introduction}
With the rapid development of Autonomous Underwater Vehicles (AUV) and Remote Operated Vehicles (ROV) \cite{zhaoConvolutionalNeuralNetwork2021}, the demands for underwater exploration also go up promptly\cite{congUnderwaterRobotSensing2021}, such as underwater scene analysis \cite{jianUnderwaterImageProcessing2021}, marine environmental surveillance\cite{baukKeyFeaturesAutonomous2021}, underwater creatures probe \cite{liUnderwaterBiologicalDetection2022}. 

Recently, underwater imaging has become the most effective way to discover the unlimited treasure troves hidden in the ocean. However, the quality of the underwater images is usually degraded via the light absorption and scattering during the propagation of light due to the particles in underwater environments \cite{chenUnderwaterImageEnhancement2021}. Besides, different wavelengths of light also affect the imaging quality. Color casts, color artifacts, and blurred details are quite common even in the high-end camera due to the poor underwater imaging environment. 

Underwater Image Enhancement (UIE) technology is invented to improve the quality of the underwater image, categorized into three types in general: physical model-based \cite{pengGeneralizationDarkChannel2018}\cite{wangNaturalnessPreservedEnhancement2013}\cite{bermanUnderwaterSingleImage2020}\cite{fuTwostepApproachSingle2017}\cite{liSingleUnderwaterImage2016}, visual prior-based\cite{ancutiEnhancingUnderwaterImages2012}\cite{iqbalEnhancingLowQuality2010}\cite{ghaniUnderwaterImageQuality2015}\cite{ancutiColorBalanceFusion2017}\cite{ancutiColorChannelCompensation2019}, and deep learning-based methods \cite{liUnderwaterImageEnhancement2019}\cite{liUnderwaterScenePrior2020}\cite{liUnderwaterImageEnhancement2021}\cite{pengUshapeTransformerUnderwater2021}\cite{dudhaneEndtoendNetworkImage2020}. Physical model-based UIE methods model the inverse process of image degradation via some prior, whose parameters require estimation. Nevertheless, the most significant problem for physical model-based UIE methods is the inability to reverse the unknown physical process, which determines that the existing physical models cannot yet generalize. It relies on specific parameters estimation, which is hardly adaptative to the dynamic underwater environment. Instead of referring to the physical models, visual prior-based UIE methods directly reconstruct the latent underwater image by adjusting image pixel values from contrast, brightness, and saturation \cite{akkaynakWhatSpaceAttenuation2017}. Without modeling any physical process, visual prior-based UIE methods result in artificial colors and excessive enhancement.

Most recently, deep learning-based methods exhibited outstanding performance. The deep learning-based UIE methods are mostly centered on the end-to-end model. As the most classic network in computing vision, the Convolutional Neural Network (CNN) \cite{wangDeepCNNMethod2017}\cite{wangUIEC2NetCNNbased2021}\cite{lyuEfficientLearningbasedMethod2022} is employed for UIE. Although depth-wise convolution can achieve high accuracy efficiently, it is tough for convolution to model the global long-range relation since its intrinsic locality is crucial to many vision tasks \cite{tanEfficientnetv2SmallerModels2021}. To address the problem, the Transformer \cite{vaswaniAttentionAllYou2017} applied self-attention modules to extend the ability and rapidly became popular in natural language processing (NLP). Furthermore, the Vision Transformer (VIT) \cite{dosovitskiyImageWorth16x162020} firstly applied the Transformer to computer vision with completely abandoning convolution. Then, the Swin Transformer \cite{liuSwinTransformerHierarchical2021} introduced shifted windows to overcome the disadvantages of ViT that destroyed the original neighborhood structure. With the popularity of the Transformer, however, the CNN seems to be overshadowed by Transformer. Many networks substituted the Convolution Layer designed to extract features with the Transformer Block. Nevertheless, revisiting the design of self-attention in the Transformer shows that the generalization of the Transformer can be worse than convolutional networks due to the lack of the right inductive bias \cite{dosovitskiyImageWorth16x162020}. Based on the description above, reintroducing the convolution into the Transformer in position is the road to combining the advantages of both.

As an end-to-end model, U-Net \cite{ronnebergerUnetConvolutionalNetworks2015}, another important architecture, can be trained to obtain excellent performance with few data. U-Net included a skipping path to capture context and a symmetric expanding path for precise localization. CNN is usually embedded into the symmetric expanding path to cooperate with U-Net. However, solely relying on CNN limits the performance improvement, while the attempt of U-Net with Transformer reported is possible to break through the limitation.

To address these problems, a novel U-Net based Reinforced Swin-Convs Transformer for Underwater Image Enhancement method (URSCT-UIE) is proposed, which is piled up by Encoder, Bottleneck, and Decoder for  capturing the global dependency and local context simultaneously. It fleetly reaches the converge, allowing the flexibility to complex and dynamic underwater environments. Besides, with the focus of URSCT intensified on global and local, we constructed a general loss function to increase the attention to textures while keeping the overall restoring consequence in equal measure.The main contributions of this paper can be summarized as follows:
\begin{itemize}
	
	\item With the deficiency of U-Net based on pure convolutions, the Swin Transformer is embedded into U-Net for improving the ability to capture the global dependency which is indispensable in UIE tasks.
	
	\item The pure Swin Transformer is normally weaker than convolutions in capturing the local attention. Thus, we provide an ingenious manner for the fusion of convolutions and the core attention mechanism of Swin Transformer to build a Reinforced Swin-Convs Transformer Block (RSCTB). 
	
	\item The proposed URSCT-UIE achieves state-of-the-art performance (SOTA) in terms of full-reference quantitative metrics on several recent benchmarks.
	
\end{itemize}

\section{The Proposed Method}
\label{sec:METHOD}
\subsection{Network Architecture}
\begin{figure}[!t]
	\centering
	\includegraphics[width=5.3in]{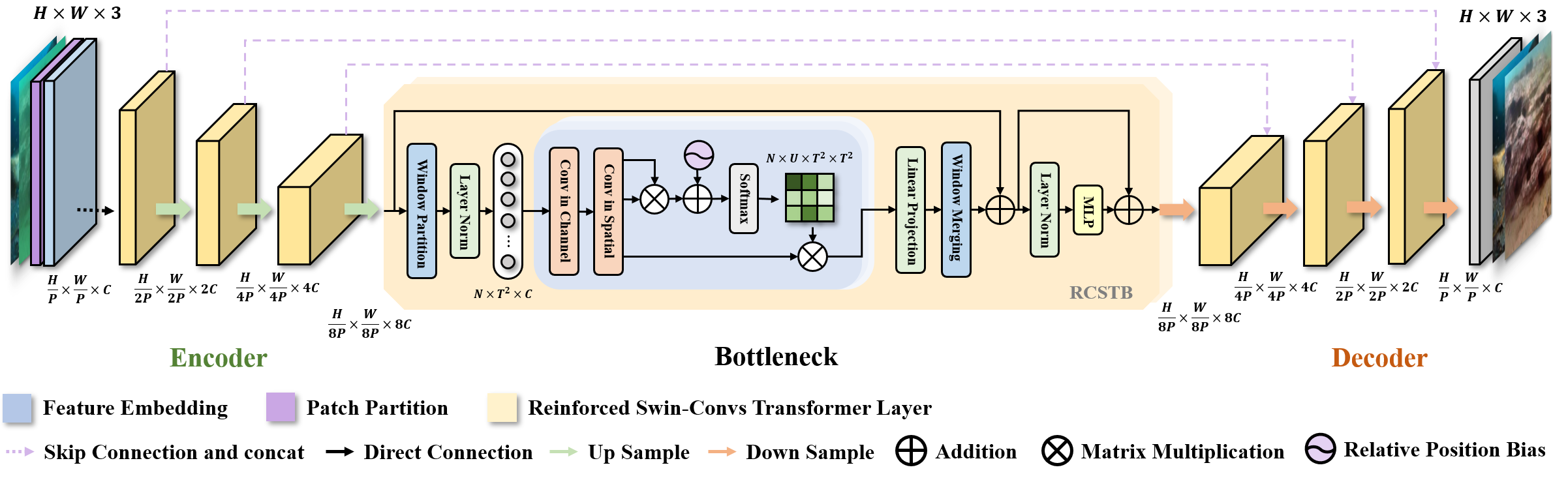}
	\caption{The architecture of the proposed network, which consist of three parts: \textbf{Encoder}, \textbf{Bottleneck}, and \textbf{Decoder}. The Encoder is used to map input into deeper feature space while the Decoder is utilized to reconstruct the image from feature space. And the Bottleneck is allowed to learn useful compression of features.}
	\label{fig:1}
\end{figure}

The overall architecture of the proposed URSCT-UIE is depicted in \autoref{fig:1}, which can be divided into three parts: \textbf{Encoder}, \textbf{Bottleneck}, and \textbf{Decoder}. The Encoder maps the given input to deeper feature space and reduces the spatial dimensions while increasing the channels. The Bottleneck locates at the bottom of the network and forces to learn the ultimately useful compression of features while maintaining feature space dimensions. The Decoder reconstructs the image from feature space and increases the spatial dimensions while reducing the channels. 

\textbf{Encoder Part}. Given a raw underwater image which resized into $\boldsymbol{X}_{in}\in \mathbb{R} ^{H\times W\times 3}$, it firstly partitioned into patches $\boldsymbol{X}_{patch}\in \mathbb{R} ^{\frac{H}{P}\times \frac{W}{P}\times 3}$, where $P$ is the size of the patch. Then, a linear projection is employed to obtain a shallow embedded feature $\boldsymbol{F}_s\in \mathbb{R} ^{\frac{H}{P}\times \frac{W}{P}\times C}$, where $C$ is the dimension of shallow features. The shallow embedded feature will be fed into a Reinforced Swin-Convs Transformer Layer to obtain $\boldsymbol{F}\in \mathbb{R} ^{\frac{H}{P}\times \frac{W}{P}\times C}$, and then transformed into deep features $\boldsymbol{F}_d\in \mathbb{R} ^{\frac{H}{2P}\times \frac{W}{2P}\times 2C}$ with down-sample. The down-sample doubles the features and twice reduces $H,W$, which preserves the structural integrity of the image to reduce distortion enormously.

\textbf{Bottleneck Part}. When the network is deep, it is meaningless to extract more features where the network falls into a so-called bottleneck. At this point, we perform feature extraction again, whose purpose is to force the network to integrate the useful information from the features to achieve the compression of features instead of extracting more features. Hence, the part is used to strengthen the global dependency. After three deep feature extractions, we put $\boldsymbol{F}_d\in \mathbb{R} ^{\frac{H}{8P}\times \frac{W}{8P}\times 8C}$ into which replaces the bottleneck part in the original U-Net. The output shape $\boldsymbol{F}_g$ will be kept since this module merely consists of a Reinforced Swin-Convs Transformer Layer to reinforce the attention in global information and catch some severely degraded parts not restored before. Then the tensor $\boldsymbol{F}_g\in \mathbb{R} ^{\frac{H}{8P}\times \frac{W}{8P}\times 8C}$ should pass through up-sample to obtain $\boldsymbol{F}_u\in \mathbb{R} ^{\frac{H}{4P}\times \frac{W}{4P}\times 4C}$ before Decoder.

\textbf{Decoder Part}. This part is utilized for the image reconstruction. Being similar to deep feature extraction, apart from the Reinforced Swin-Convs Transformer Layer retained, patch merging and down-sample are replaced by up-sample, respectively. After three reconstructions, features are reduced to $\boldsymbol{F}_u\in \mathbb{R} ^{H\times W\times C}$. Here, we set $P=2$. If the size of the patch is set to 4 or others, the last up-sample need to be distinctively defined. In the end, the underwater image will be restored to $\boldsymbol{\hat{X}}\in \mathbb{R} ^{H\times W\times 3}$ via a convolution projecting embedded features to RGB channels.

\subsection{Reinforced Swin-Convs Transformer Block}
It is not adequate that capturing the local attention merely relies on STB, which is mainly responsible for global dependency. Given that STB will divide the image into several small windows, convolutions may capture more local attention in every window. It is possible that the fusion of STB and convolutions can interflow and mutual learning. 

\begin{figure}[h]
	\centering
	\subfloat[\label{fig:a}]{
		\includegraphics[scale=0.36]{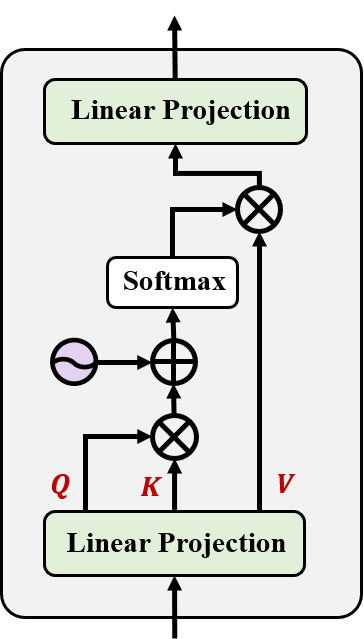}}
	\subfloat[\label{fig:c}]{
		\includegraphics[scale=0.3125]{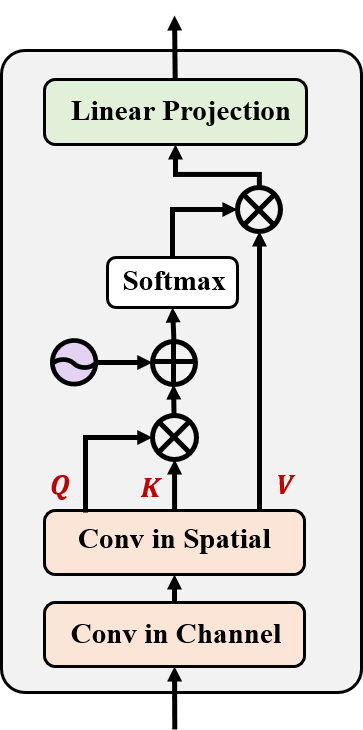}}
	\subfloat[\label{fig:e}]{
		\includegraphics[scale=0.31]{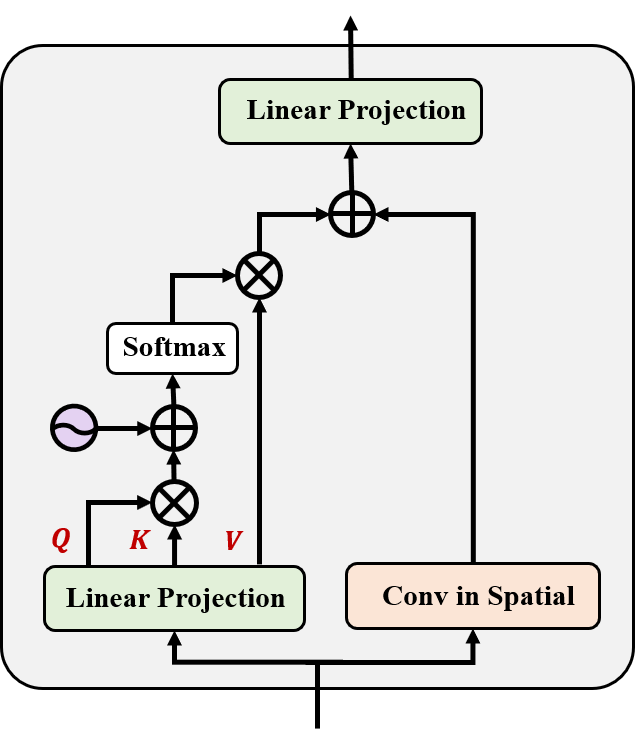}}
	\caption{ The different manners to exploit convolutions in the attention calculation module of STB. The detailed description of the components, see below. (a) The original manner. (b) The improved manner I (proposed). (c) The improved manner II.}
	\label{fig:2}
\end{figure}

The inclusion of convolution must follow the constraint of not changing the core attention mechanism. We provide and compare two different manners to exploit convolutions in the attention calculation module of STB, as shown in \autoref{fig:2}. The performance of these methods will be illustrated in the Ablation Study. The first manner, the improvement used in our URSCT, skillfully takes the coincidence that the way to generate $Q$, $K$, $V$ reported in Section I to introduce convolutions, which simultaneously reinforce the local attention in the channel and spatial. Another manner directly constructs another branch to unite convolutions for strengthening the local attention only spatial-wise.

Before introducing the proposed RSCTB, it is necessary to figure out the relationship between the Reinforced Swin-Convs Transformer Layer and the Reinforced Swin-Convs Transformer Block. In the presentation of network architecture above, the Reinforced Swin-Convs Transformer Layer contains even times consecutive RSCTBs.
%

The original STB keeps the roughly same architecture as Transformer Block, except in several added modules for handling windows and the standard multi-head self-attention (MSA) module replaced by a similar module multi-head self-attention (W-MSA) based on windows, while SW-MSA based on shifted windows.

As for our RSCTB, obviously, it introduces the convolution as the name shows. A small convolution kernel is often used in the image enhancement task to avoid a too sizeable receptive field \cite{shangPerceptualExtremeSuperresolution2020}. We hope to enhance local attention in each patch. Thus, in the W-MSA module, we use two Convolutions instead of a linear layer to respectively reinforce in the channel and spatial. In channel, it is achieved by a 1x1 convolution to triple the channels, which is like a linear layer; In spatial, it is enhanced by a 3x3 convolution in channel-wise. 


For an input tensor $\boldsymbol{Z}^l\in \mathbb{R} ^{H\times W\times C}$, the Window Partition (WP) module splits it into multiple windows $\boldsymbol{Y}^l\in \mathbb{R} ^{N\times T^2\times C}$, where $N$ is the number of the windows and $T$ is the window size. With layer normalization, we obtain $\boldsymbol{Y}^{l-1}\in \mathbb{R} ^{N\times T^2\times C}$. Then, normalized features will be fed into W-MSA. Firstly, Query $Q\in \mathbb{R} ^{N\times U\times T^2\times \frac{C}{U}}$, Key $K\in \mathbb{R} ^{N\times U\times T^2\times \frac{C}{U}}$, and Value $V\in \mathbb{R} ^{N\times U\times T^2\times \frac{C}{U}}$ are generated by two convolutions $\mathrm{Con}^{\mathrm{s}}$ and $\mathrm{Con}^{\mathrm{v}}$improved above, where $U$ is the head num. 
With the relative position $\boldsymbol{B}\in \mathbb{R} ^{U\times T^2\times T^2}$ as a bias parameter which follows \cite{raffelExploringLimitsTransfer2019}, the self-attention in $\mathbb{R} ^{N\times U\times T^2\times T^2}$ is defined as:
\[
\mathrm{Attention}\left( \boldsymbol{Q},\boldsymbol{K},\boldsymbol{V} \right) =\mathrm{SoftMax}\left( \frac{\boldsymbol{QK}'}{\sqrt{C}}+\boldsymbol{B} \right) \boldsymbol{V}
\]
If RSCTB is located in Decoder, mask mechanism should be applied next. At the end, we concatenate all heads’ attention via a linear projection to obtain the output of the Conv-W-MSA in $
\mathbb{R} ^{N\times T^2\times C}$.

Likewise, the Window Merging (WM) module reverses the process above to obtain $\boldsymbol{Z}^l\in \mathbb{R} ^{H\times W\times C}$. After residual connection and layer normalization, it is Multi-Layer Perceptron (MLP) with GELU that plays a role as feedforward. At the end, the output of RSCTB $\boldsymbol{Z}^l$is accessed with residual connection:
\[
\boldsymbol{Z}^l=\mathrm{MLP}\left( \mathrm{LN}\left( \boldsymbol{\hat{Z}}^l \right) \right) +\boldsymbol{\hat{Z}}^l
\]

The RSCTB is consecutively paired. In another RSCTB, merely Conv-W-MSA is replaced by Conv-SW-MSA, where only Cyclic Shift (CS) is extremely introduced which is employed to shift the window in a particular way reported in \cite{liuSwinTransformerHierarchical2021}.
\[
\mathrm{Conv}\text{-}\mathrm{SW}\text{-}\mathrm{MSA}\left( \boldsymbol{Y}^l \right) =\mathrm{CS}\left( \mathrm{Conv}\text{-} \mathrm{W}\text{-} \mathrm{MSA}\left( \boldsymbol{Y}^l \right) \right) 
\]

In summary, the consecutive paired CSTBs are computed as the following:
\begin{flalign*}
&& \boldsymbol{Y}^{l-1}&=\mathrm{LN}\left( \mathrm{WP}\left( \boldsymbol{Z}^{l-1} \right) \right) & \boldsymbol{Y}^l&=\mathrm{LN}\left( \mathrm{WP}\left( \boldsymbol{Z}^l \right) \right) \\
&& \boldsymbol{\hat{Z}}^l&=\mathrm{WM}\left( \mathrm{Conv}\text{-}\mathrm{W}\text{-}\mathrm{MSA}\left( \boldsymbol{Y}^{l-1} \right) \right) +\boldsymbol{Z}^{l-1} & \boldsymbol{\hat{Z}}^{l+1}&=\mathrm{WM}\left( \mathrm{Conv}\text{-}\mathrm{SW}\text{-}\mathrm{MSA}\left( \boldsymbol{Y}^l \right) \right) +\boldsymbol{Z}^l \\
&& \boldsymbol{Z}^l&=\mathrm{MLP}\left( \mathrm{LN}\left( \boldsymbol{\hat{Z}}^l \right) \right) +\boldsymbol{\hat{Z}}^l & \boldsymbol{Z}^{l+1}&=\mathrm{MLP}\left( \mathrm{LN}\left( \boldsymbol{\hat{Z}}^{l+1} \right) \right) +\boldsymbol{\hat{Z}}^{l+1}
\end{flalign*}


As the core of the proposed URSCT-UIE, the RCSTB only improves the W-MSA module of the original Swin Transformer Block, achieving the reinforcement of the local attention in the channel and the spatial. The introduction of the convolutions has shown excellent results on some challenging images (see \hyperref[sec:AbS]{\textbf{Ablation Study}}).

\subsection{Loss Function}

\textbf{Charbonnier loss}. The L2 loss which means maximizing the log-likelihood of Gaussian, usually causes the image to be blurry in reconstruction \cite{isolaImagetoimageTranslationConditional2017}. Thus, we first use the robust Charbonnier loss \cite{bruhnLucasKanadeMeets2005}, a differentiable variant of the L1 norm, to minimize the loss between restored and reference images.  Given the restored image $\boldsymbol{\hat{X}}$ and the reference image $\boldsymbol{X}$, the Charbonnier loss between them is defined as follows:
\[
\mathcal{L} _{\mathrm{C}}=\mathbb{E} _{\boldsymbol{\hat{X}}\sim P\left( r \right) , \boldsymbol{X}\sim P\left( g \right)}\sqrt{\left( \boldsymbol{\hat{X}}-\boldsymbol{X} \right) ^2+\epsilon ^2}
\]
where $P\left( r \right) $ and $P\left( g \right) $ are the distribution of the restored images $\boldsymbol{\hat{X}}$ and the real images $\boldsymbol{X}$, respectively. Besides, we empirically set $\epsilon $ to 1e – 3.

\textbf{Gradient loss}. The Charbonnier loss only obtains low-frequency information like the L1 loss and the gradient loss \cite{mathieuDeepMultiscaleVideo2015} via imposing a second-order restriction to capture the high-frequency information, which is beneficial to sharpen the edge of the enhanced image. Let $\boldsymbol{\hat{G}}$ and $\boldsymbol{G}$ denotes the gradient map of $\boldsymbol{\hat{X}}$ and $\boldsymbol{X}$, respectively, L1 gradient loss is expressed as follows:
\[
\mathcal{L} _{\mathrm{gd}}=\mathbb{E} _{\boldsymbol{\hat{G}}\sim Q\left( r \right) , \boldsymbol{G}\sim Q\left( g \right)}\left\| \boldsymbol{\hat{G}}-\boldsymbol{G} \right\| _1
\]
where $Q\left( r \right) $ and $Q\left( g \right) $ are the distribution of $\boldsymbol{\hat{G}}$ and $\boldsymbol{G}$, respectively.

\textbf{MS-SSIM loss}. The Multi-Scale Structural Similarity (MS-SSIM) \cite{wangMultiscaleStructuralSimilarity2003} is a structural similarity image quality paradigm, which provides more flexibility than Single-Scale Structural Similarity (SSIM) \cite{wangImageQualityAssessment2004} in incorporating the variations of image resolution and viewing conditions \cite{wangMultiscaleStructuralSimilarity2003}. Thus, the corresponding loss is defined as follows:
\[
\mathcal{L} _{\mathrm{M}}\left( \boldsymbol{\hat{X}},\boldsymbol{X} \right) =1-\mathrm{MS}\-\mathrm{SSIM}\left( \boldsymbol{\hat{X}},\boldsymbol{X} \right) 
\]

Finally, the total loss function is expressed as:
\[
\mathcal{L} _{\mathrm{sum}}\,\,=\,\,w_1\mathcal{L} _{\mathrm{C}}+w_2\mathcal{L} _{\mathrm{gd}}+w_3\mathcal{L} _{\mathrm{M}}
\]

The hyperparameters determine the balance between the overall performance and the local texture details. To maintain the convergence rate simultaneously, $w_1$, $w_2$, and $w_3$ are set empirically as $1$, $1$, and $2$ according to the experiments.

\section{Experiments and Analysis}
\label{sec:EXP}
\subsection{Training Details}
	\textbf{Implement Details}. The proposed URSCT-UIE is implemented by Pytorch 1.11.0 with an NVIDIA RTX 3090 GPU without pre-trained networks. Adam optimizer with the initial learning rate 5e-4 and $\beta \in \left( 0.9,0.999 \right) $ is utilized for the training, processing 800 epochs with the batch size 8. Besides, the cosine annealing learning rate decay strategy \cite{loshchilovDecoupledWeightDecay2017} is synergized for training with the warmup epoch 3.
	
	\textbf{Hyperparameters Details}. The hyperparameters settings of our networks are as follows: the window size is 8, the patch size is 2, the depth of each Reinforced Swin-Convs Transformer Layer is 8, the head num of each RSCTB is also 8, the embedding dimension is 32, the dropping ratio in skip path is 0.1, the scale factor in $Q$, $K$ is 8, and the ratio multiplied in MLP is 4.

\subsection{Datasets and Evaluation Metrics}
\textbf{Datasets}. In this paper, four datasets are used:
\begin{enumerate} 
	\item \textbf{UIEB dataset}. The underwater Image Enhancement Benchmark (UIEB) dataset \cite{liUnderwaterImageEnhancement2019} includes 950 real-world underwater images grouped into two subsets: 890 pairs of raw underwater images with the corresponding high-quality reference images and 60 challenging images without reference.
	
	\item \textbf{LSUI dataset}. The large-scale underwater image (LSUI) dataset \cite{pengUshapeTransformerUnderwater2021} includes 5004 natural underwater images pairs. It has better visual quality reference images and involves more diverse underwater environments than the existing underwater datasets.
	
	\item \textbf{SQUID dataset}. The Stereo Quantitative Underwater Image Dataset (SQUID) \cite{bermanUnderwaterSingleImage2020} contains 57 stereo pairs from four different sites in Israel with varying water properties.
	
	\item \textbf{UFO-120 dataset}. The UFO-120 dataset \cite{islamSimultaneousEnhancementSuperresolution2020} contains 1500 synthetic images for training and 120 synthetic images for testing, simultaneously for enhancement and super-resolution. The high-resolution ground truth is collected from oceanic explorations in multiple locations, while the distorted images are generated by CycleGAN trained on unpaired natural data and then follow Gaussian blurring and bicubic down-sampling.
\end{enumerate}

\textbf{Full-Reference Evaluation}. To quantitatively compare the restored images with the paired reference images offered on the dataset, we conduct the evaluation using PSNR  \cite{korhonenPeakSignaltonoiseRatio2012} and SSIM \cite{wangImageQualityAssessment2004} metrics, which reflect the proximity to the reference. The higher values of PSNR and SSIM, the more similar structure between images.

\textbf{No-Reference Evaluation}. It is a hard mission that evaluates the UIEs comprehensively and fairly for non-reference testing datasets. UCIQE \cite{yangUnderwaterColorImage2015} and UIQM \cite{panettaHumanvisualsysteminspiredUnderwaterImage2015} are employed. UCIQE focus on Color density, saturation, and contrast, while UIQM concentrates on underwater image color (UICM), underwater image sharpness measure (UISM), and underwater image contrast measure (UIConM) \cite{chenMFFNUnderwaterSensing2021}. A higher UCIQE or UIQM score indicates a better visual perception.

\subsection{Qualitative Comparison}
In this subsection, we first show the qualitative results on full-reference datasets. We conducted comprehensive experiments on the UIEB dataset. The testing data is the remaining 90 images from the UIEB dataset, denoted as \textbf{Test-U}. Furthermore, the rest of the 800 paired images are utilized as training data, denoted as \textbf{Train-U}.

\begin{figure}[h]
	\centering
	\subfloat[\label{fig:a}]{
		\includegraphics[scale=0.25]{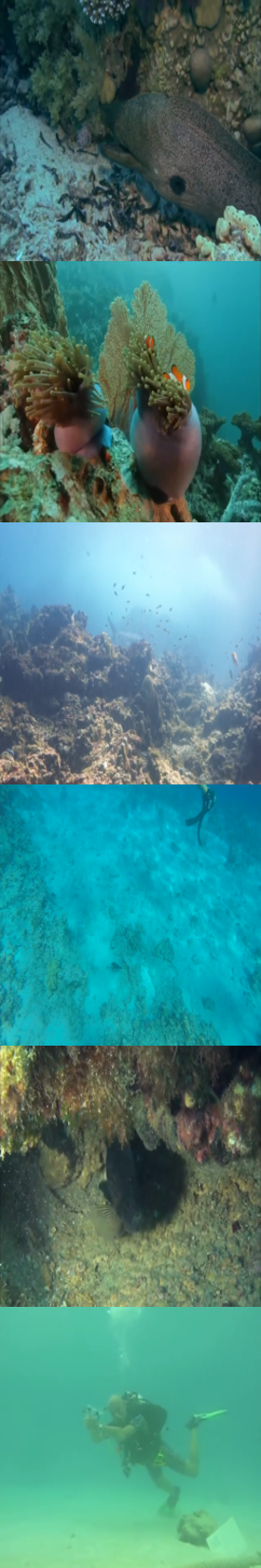}}
	\subfloat[\label{fig:c}]{
		\includegraphics[scale=0.25]{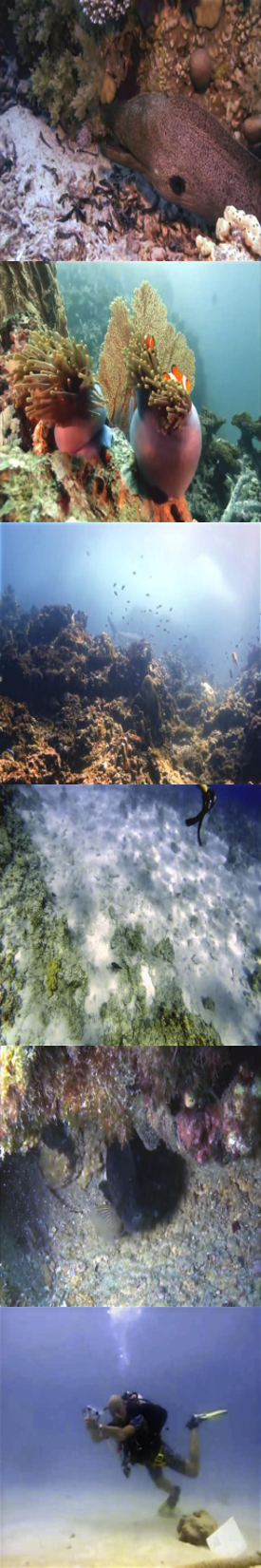}}
	\subfloat[\label{fig:e}]{
		\includegraphics[scale=0.25]{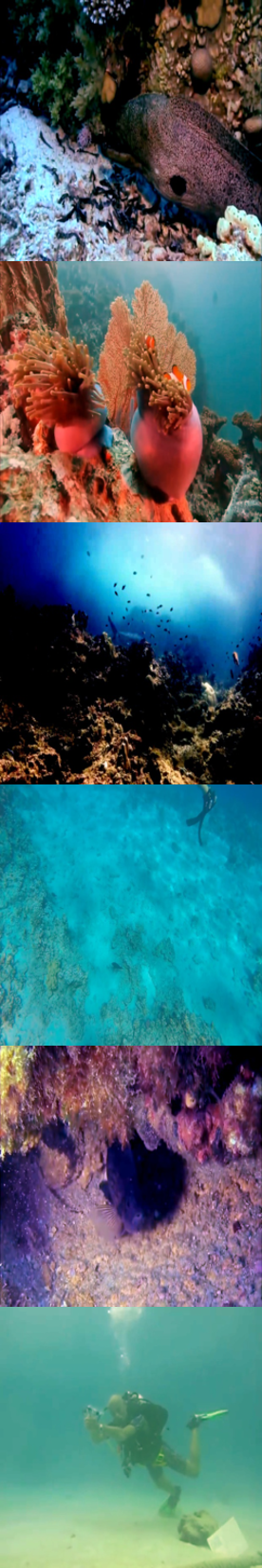}}
	\subfloat[\label{fig:e}]{
		\includegraphics[scale=0.25]{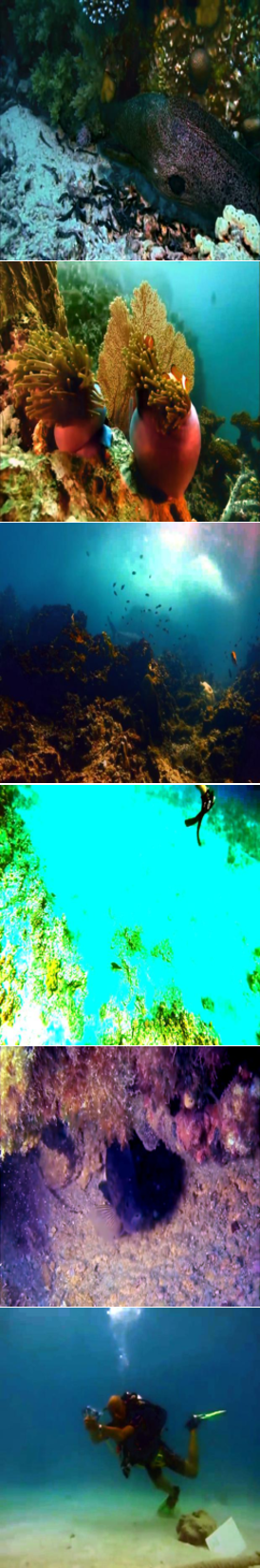}}
	\subfloat[\label{fig:e}]{
		\includegraphics[scale=0.25]{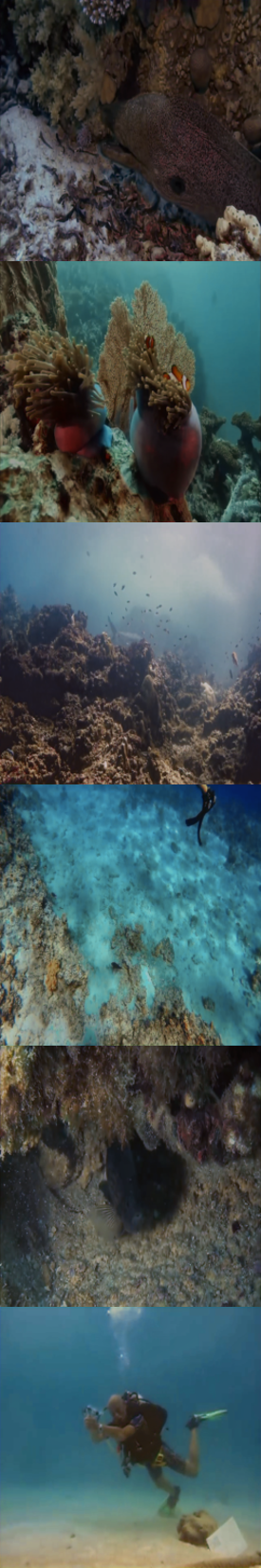}}
	\subfloat[\label{fig:e}]{
		\includegraphics[scale=0.25]{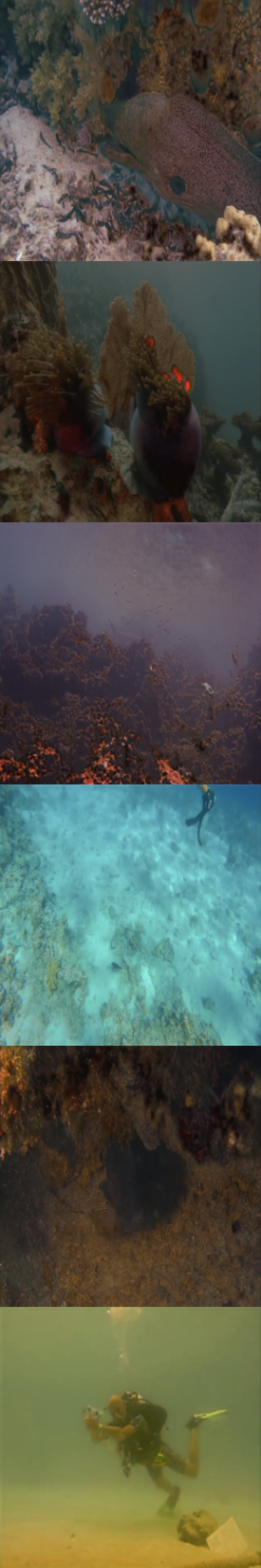}}
	\subfloat[\label{fig:e}]{
		\includegraphics[scale=0.25]{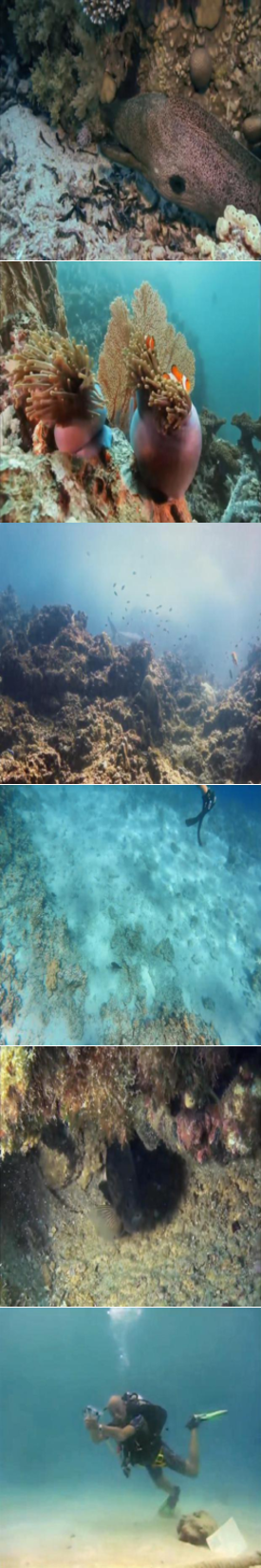}}
	\subfloat[\label{fig:e}]{
		\includegraphics[scale=0.25]{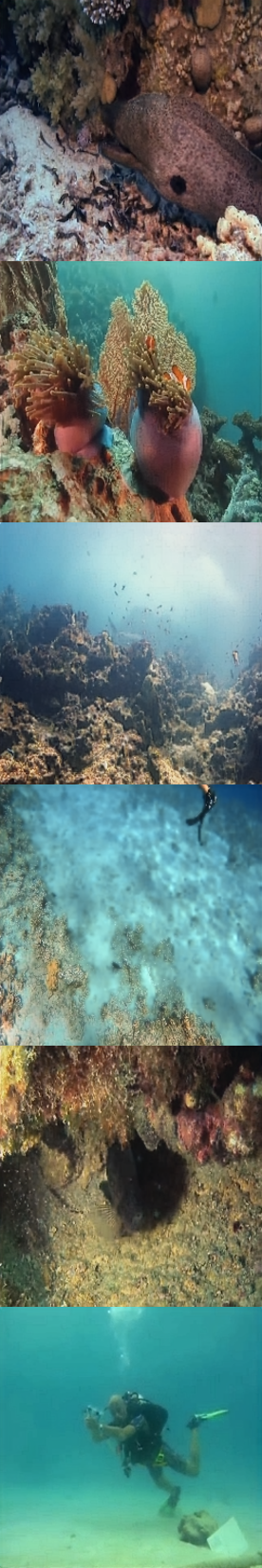}}
	\subfloat[\label{fig:e}]{
		\includegraphics[scale=0.25]{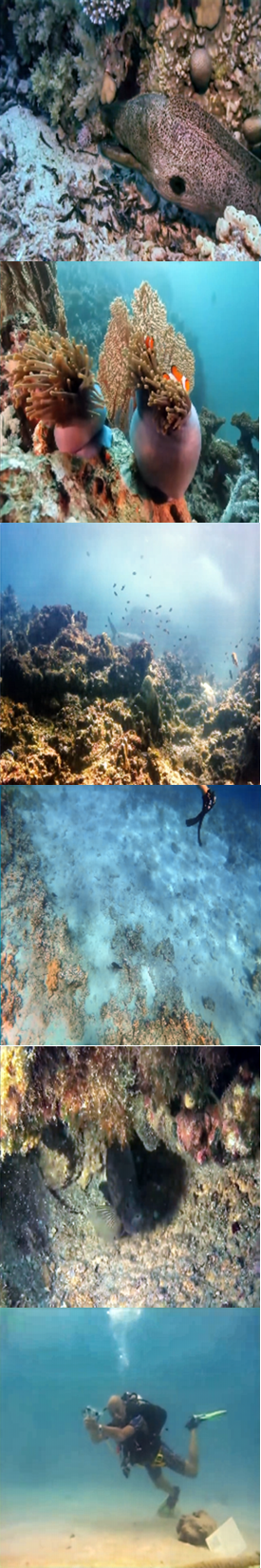}}
	\subfloat[\label{fig:e}]{
		\includegraphics[scale=0.25]{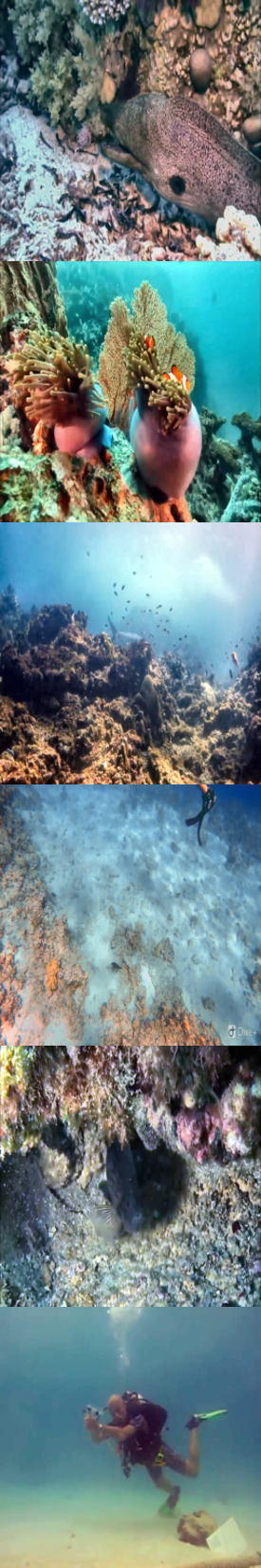}}
	
	\caption{The Visual comparisons on underwater images sampled from Test-U. (a) RAWS. (b) Fusion \cite{ancutiEnhancingUnderwaterImages2012}. (c) IBLA \cite{wangNaturalnessPreservedEnhancement2013}. (d) HL \cite{bermanUnderwaterSingleImage2020}. (e) WaterNet \cite{liUnderwaterImageEnhancement2019}. (f) UWCNN-typeI \cite{liUnderwaterScenePrior2020}. (g) Ucolor \cite{liUnderwaterImageEnhancement2021}. (h) U-Trans \cite{pengUshapeTransformerUnderwater2021}. (i) The proposed URSCT-UIE. (j) The ground truth.}
	\label{fig:5}
\end{figure}

The compared methods include visual prior-based UIE Fusion \cite{ancutiEnhancingUnderwaterImages2012}, physical model-based UIE IBLA \cite{wangNaturalnessPreservedEnhancement2013} and HL \cite{bermanUnderwaterSingleImage2020}, and deep learning-based UIE WaterNet \cite{liUnderwaterImageEnhancement2019}, UWCNN-typeI \cite{liUnderwaterScenePrior2020}, Ucolor \cite{liUnderwaterImageEnhancement2021}, and U-Trans \cite{pengUshapeTransformerUnderwater2021}. There are six underwater environments on \textbf{Test-U} for comparison: shadowed, texture, hazing, blueish, yellowish, and greenish scenes. The most representative images of each type are sampled, whose enhancement results of different methods are shown in \autoref{fig:5}.

We also show the qualitative results on the no-reference datasets. The comprehensive experiments are conducted on two challenging benchmarks. The first benchmark is 60 challenging images from the UIEB dataset, denoted as \textbf{Test-C60}. And another benchmark is sampled 16 representative examples from the whole SQUID dataset, denoted as \textbf{Test-S16}. Besides, the whole dataset LSUI (5004 images) is used for the training. 

\begin{figure}[h]
	\centering
	\subfloat[\label{fig:a}]{
		\includegraphics[scale=0.3]{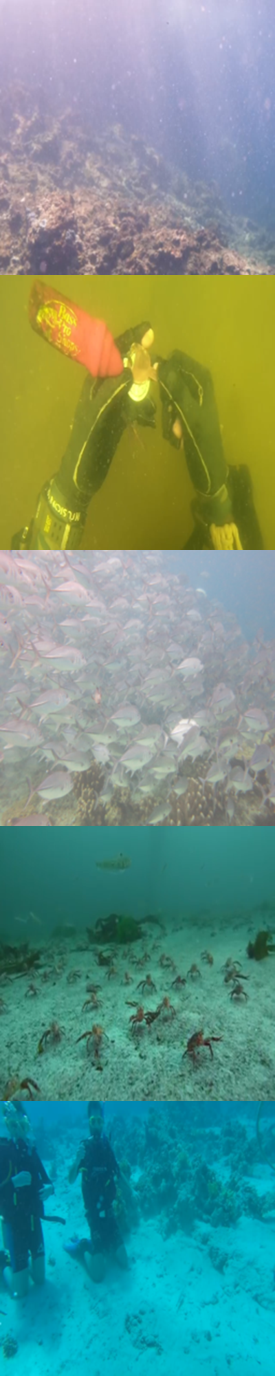}}
	\subfloat[\label{fig:c}]{
		\includegraphics[scale=0.3]{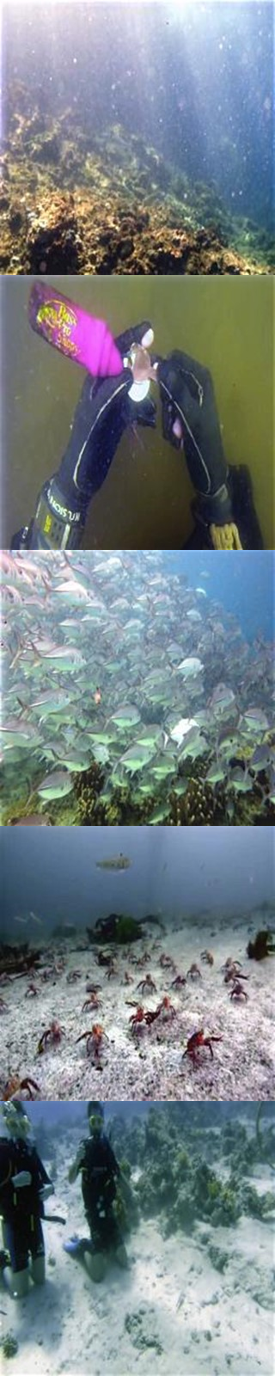}}
	\subfloat[\label{fig:e}]{
		\includegraphics[scale=0.3]{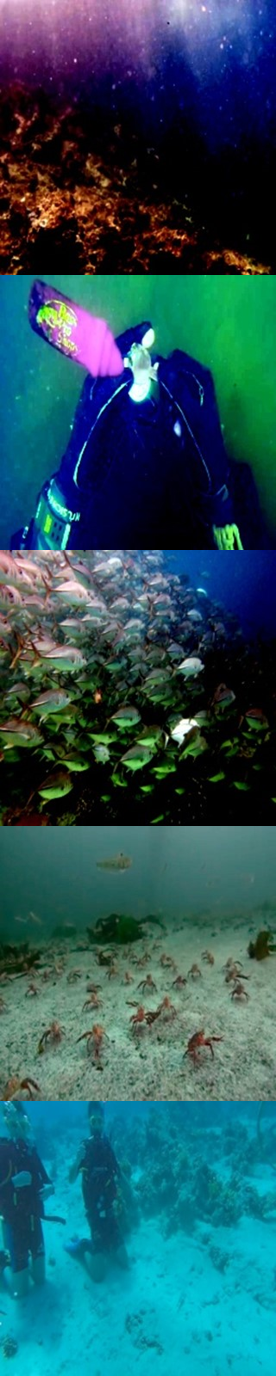}}
	\subfloat[\label{fig:e}]{
		\includegraphics[scale=0.3]{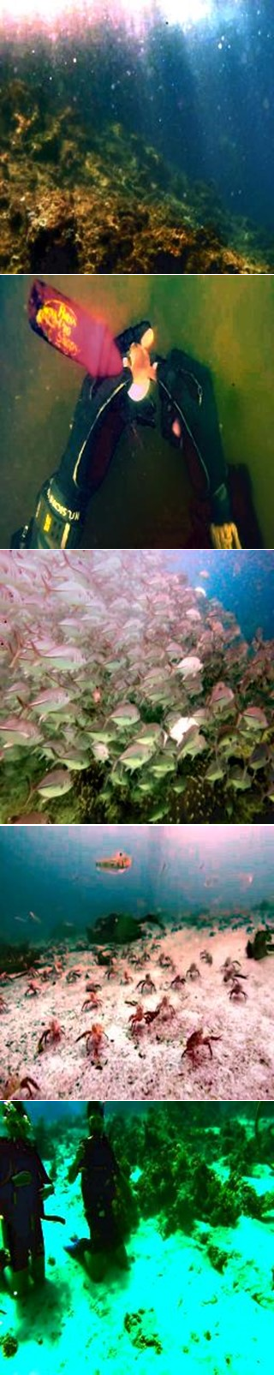}}
	\subfloat[\label{fig:e}]{
		\includegraphics[scale=0.3]{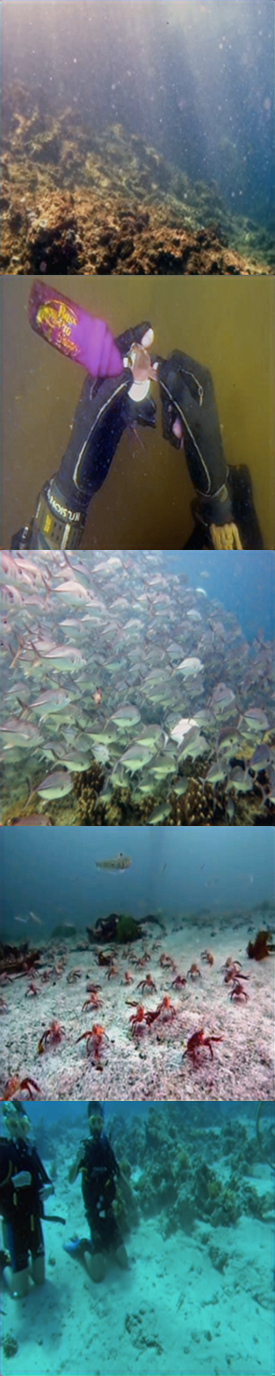}}
	\subfloat[\label{fig:e}]{
		\includegraphics[scale=0.3]{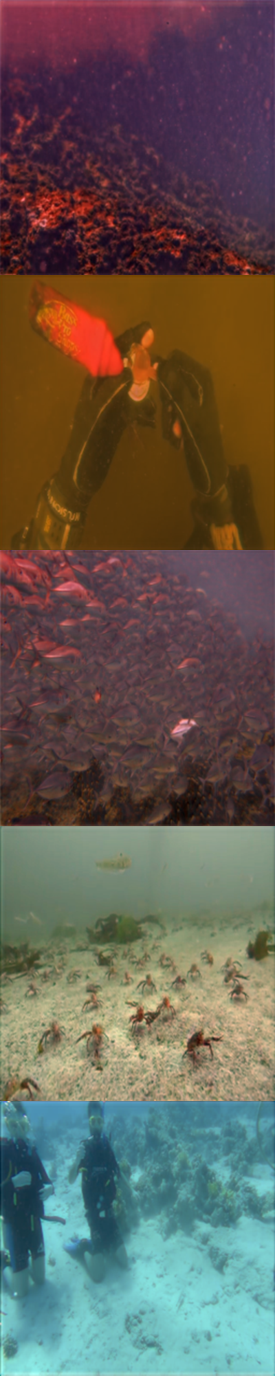}}
	\subfloat[\label{fig:e}]{
		\includegraphics[scale=0.3]{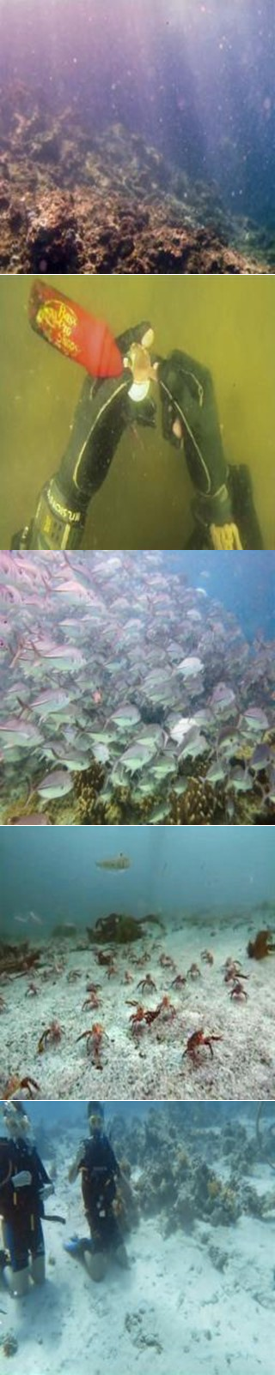}}
	\subfloat[\label{fig:e}]{
		\includegraphics[scale=0.3]{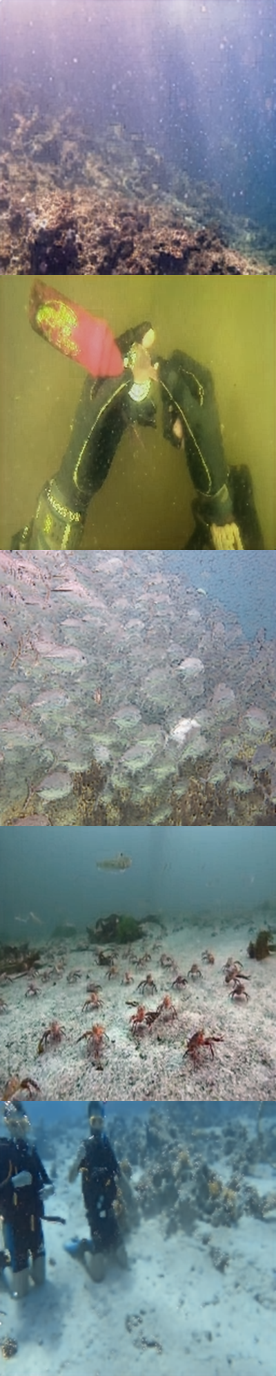}}
	\subfloat[\label{fig:e}]{
		\includegraphics[scale=0.3]{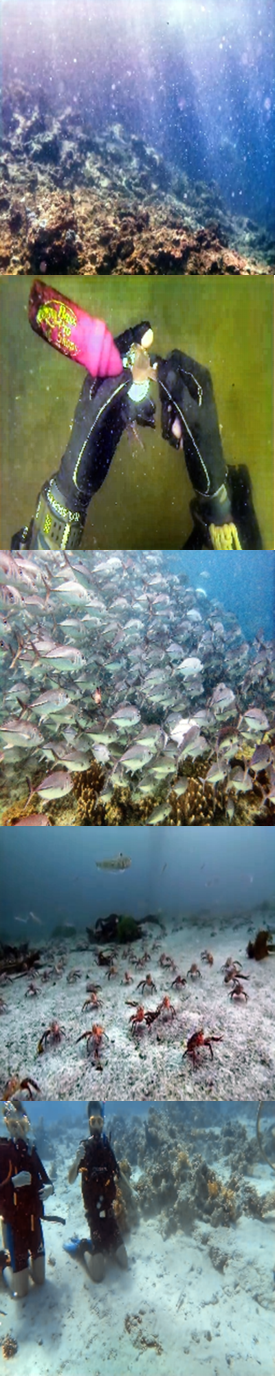}} 
	
	\caption{The enhancement results of different methods in \textbf{Test-C60}. From top to bottom are reddish, yellowish, greenish, blueish, and hazing scenes. (a) RAWS. (b) Fusion \cite{ancutiEnhancingUnderwaterImages2012}. (c) IBLA \cite{wangNaturalnessPreservedEnhancement2013}. (d) HL \cite{bermanUnderwaterSingleImage2020}. (e) WaterNet \cite{liUnderwaterImageEnhancement2019}. (f) UWCNN-typeI \cite{liUnderwaterScenePrior2020}. (g) Ucolor \cite{liUnderwaterImageEnhancement2021}. (h) U-Trans \cite{pengUshapeTransformerUnderwater2021}. (i) The proposed URSCT-UIE.}
	\label{fig:6}
\end{figure}

\textbf{Test-C60} includes five types of underwater environments: reddish, yellowish, greenish, blueish, and hazing scenes, which suffer from high backscattering and color deviations. We sampled the most representative images of each type for visual comparison. As shown in \autoref{fig:6}, while the proposed URSCT-UIE recovered the relatively realistic color and enhanced the details compared with others.


\begin{figure}[h]
	\centering
	\subfloat[\label{fig:a}]{
		\includegraphics[scale=0.3]{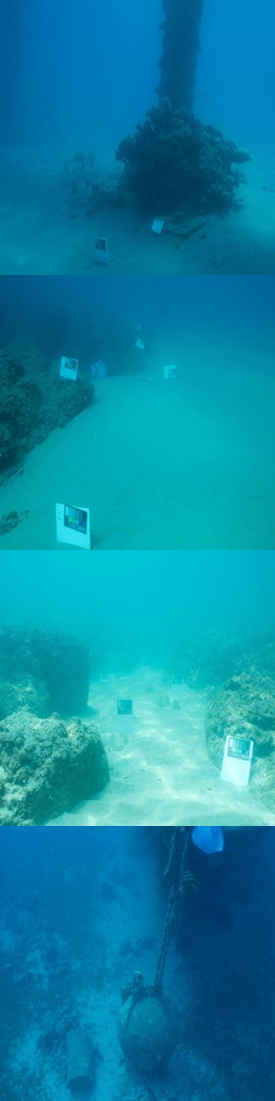}}
	\subfloat[\label{fig:c}]{
		\includegraphics[scale=0.3]{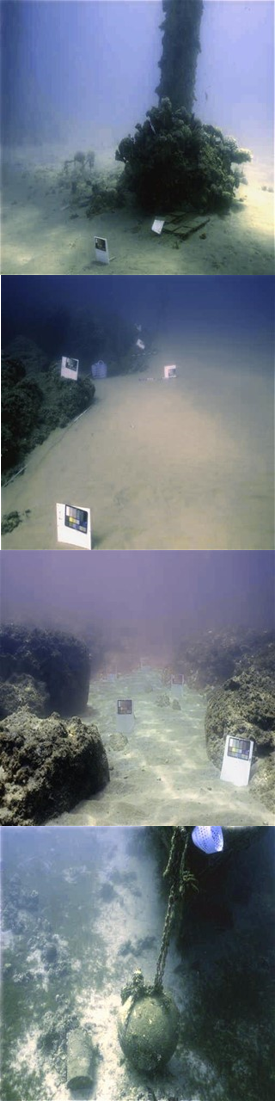}}
	\subfloat[\label{fig:e}]{
		\includegraphics[scale=0.3]{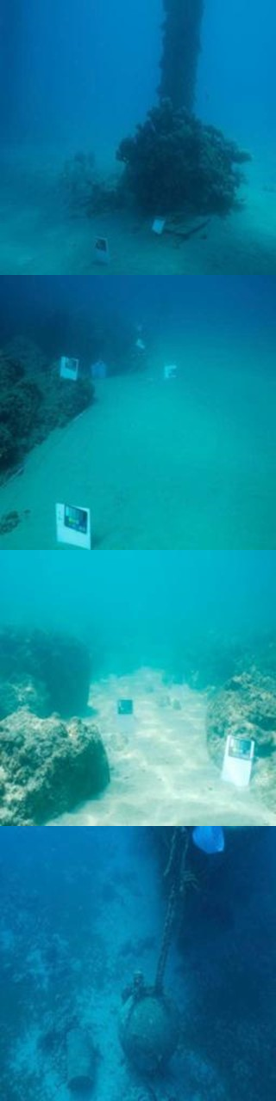}}
	\subfloat[\label{fig:e}]{
		\includegraphics[scale=0.3]{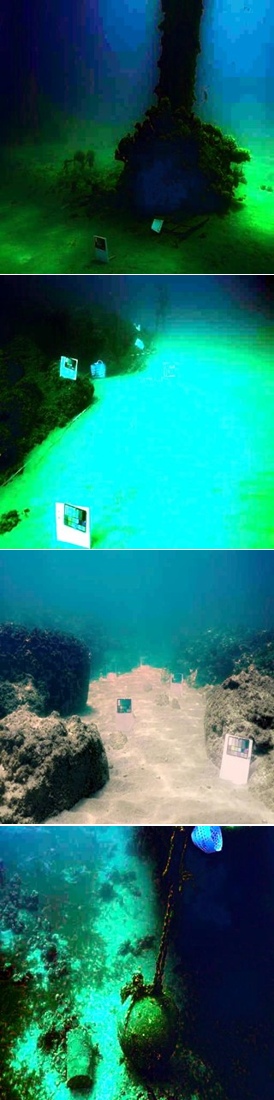}}
	\subfloat[\label{fig:e}]{
		\includegraphics[scale=0.3]{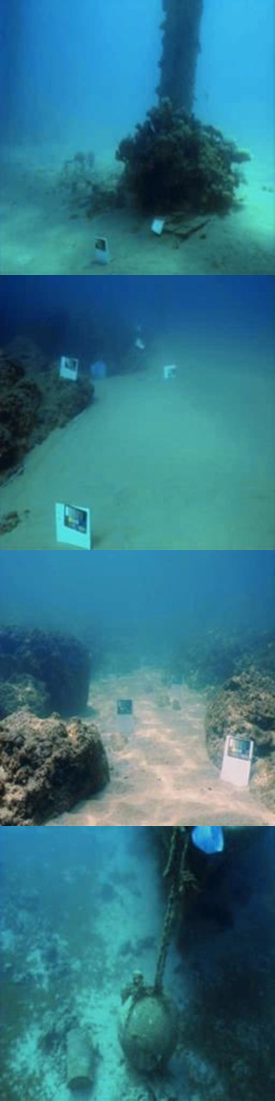}}
	\subfloat[\label{fig:e}]{
		\includegraphics[scale=0.3]{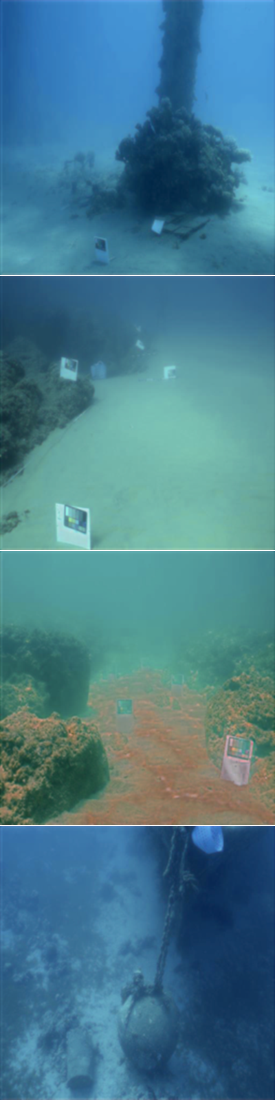}}
	\subfloat[\label{fig:e}]{
		\includegraphics[scale=0.3]{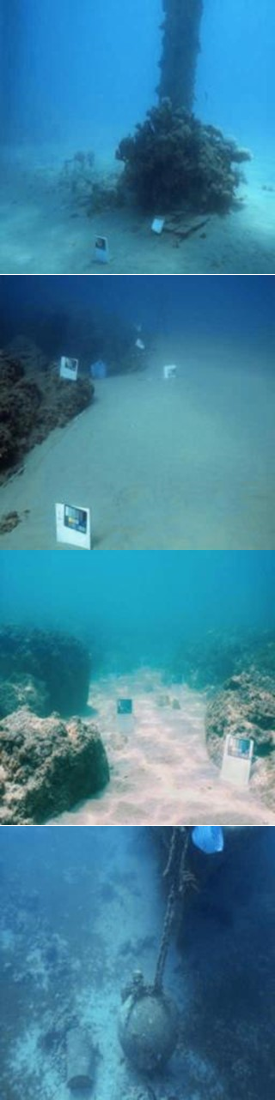}}
	\subfloat[\label{fig:e}]{
		\includegraphics[scale=0.3]{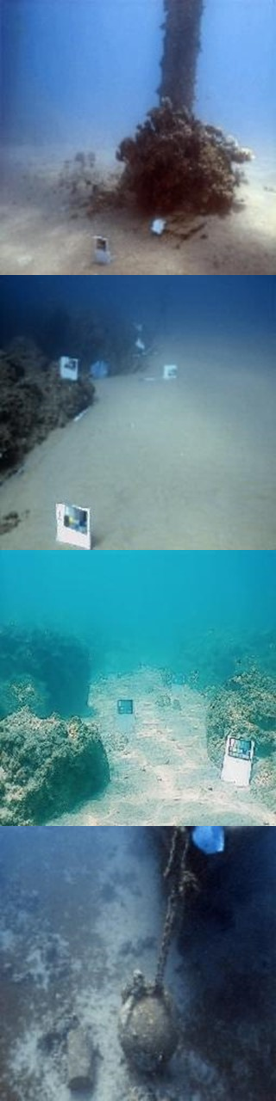}}
	\subfloat[\label{fig:e}]{
		\includegraphics[scale=0.3]{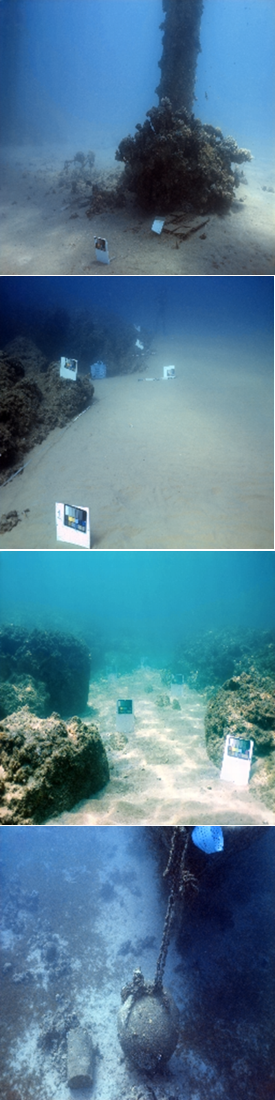}} 
	
	\caption{The enhancement results of different methods in \textbf{Test-S16}. The images from top to bottom represent Katzaa, Michmoret, Nachsholim, and Satil. (a) RAWS. (b) Fusion \cite{ancutiEnhancingUnderwaterImages2012}. (c) IBLA \cite{wangNaturalnessPreservedEnhancement2013}. (d) HL \cite{bermanUnderwaterSingleImage2020}. (e) WaterNet \cite{liUnderwaterImageEnhancement2019}. (f) UWCNN-typeI \cite{liUnderwaterScenePrior2020}. (g) Ucolor \cite{liUnderwaterImageEnhancement2021}. (h) U-Trans \cite{pengUshapeTransformerUnderwater2021}. (i) The proposed URSCT-UIE.}
	\label{fig:7}
\end{figure}

\textbf{Test-S16} contains four different dive sites in Israel: Katzaa, Michmoret, Nachsholim, and Satil. These sites represent coral reefs with 10-15 meters deep, rocky reef at 10-12 meters depth, rocky reef at 3-6 meters depth and shipwreck with 20-30 meters deep, respectively. As shown in \autoref{fig:7}, the proposed URSCT-UIE enhanced the contrast and removed the blue artifact effectively.



Besides, the qualitative comparison on the cross-dataset is also shown in \hyperref[sec:App]{\textbf{Appendix}}.

\subsection{Quantitative Comparison}
In this subsection, we first show the quantitative results on full-reference, i.e., the \textbf{Test-U} dataset. As the summarized statistical results presented in \autoref{tab:1}, our URSCT has become state-of-the-art in both PSNR and SSIM compared with the current methods. Moreover, our URSCT achieves a percentage gain of 8.3\%/6.1\% in terms of PSNR/SSIM compared with the second-best performer.
\begin{table}
	\caption{The PSNR scores and SSIM scores of different methods on \textbf{Test-U90}. The best result is in red under each case.}
	\label{tab:1}
	\centering
	\begin{tabular}{cccc}
		\toprule
	 & & \multicolumn{2}{c}{\textbf{Test-U}}   \\ \cmidrule(r){3-4}
	\multirow{-2}{*}{\textbf{Training}} & \multirow{-2}{*}{\textbf{Methods}}     & PSNR   & SSIM   \\ \midrule
&Fusion \cite{ancutiEnhancingUnderwaterImages2012}   & 19.04                & 0.82                  \\ 
&IBLA \cite{wangNaturalnessPreservedEnhancement2013}     & 15.78                & 0.73                        \\
&HL \cite{bermanUnderwaterSingleImage2020}     & 15.11                & 0.70                         \\
&WaterNet \cite{liUnderwaterImageEnhancement2019}    & 19.81                & 0.86                        \\
&UWCNN-typeI \cite{liUnderwaterScenePrior2020} & 13.76                & 0.61                        \\
&Ucolor \cite{liUnderwaterImageEnhancement2021}      & 20.78               & 0.87                        \\
&U-Trans \cite{pengUshapeTransformerUnderwater2021}       & 21.25               & 0.84                        \\
\multirow{-8}{*}{\textbf{Train-U}} &Ours    & {\color[HTML]{C00000}\textbf{22.72}} & {\color[HTML]{C00000} \textbf{0.91}} \\ \bottomrule
\end{tabular}
\end{table}

\begin{table}
	\caption{The UIQM scores and UCIQE scores of different methods on \textbf{Test-C60} and \textbf{Test-S16}. The best result is in red under each case.}
	\label{tab:1}
	\centering
	\begin{tabular}{cccccc}
		\toprule
			& & \multicolumn{2}{c}{\textbf{Test-C60}}  & \multicolumn{2}{c}{\textbf{Test-S16}}    \\ \cmidrule{3-6}
			\multirow{-2}{*}{\textbf{Training}} & \multirow{-2}{*}{\textbf{Methods}}  & UIQM  & UCIQE & UIQM  & UCIQE   \\ \midrule
		&Fusion \cite{ancutiEnhancingUnderwaterImages2012} 
		& 2.6236   & 0.5756  & 1.7863   & 0.5733  \\
		&IBLA \cite{wangNaturalnessPreservedEnhancement2013}& {\color[HTML]{C00000} \textbf{3.3281}} & 0.5687 & 1.2582  & 0.471  \\
		&HL \cite{bermanUnderwaterSingleImage2020}   & 2.6283    & {\color[HTML]{C00000} \textbf{0.6384}} & 2.1404    & {\color[HTML]{C00000} \textbf{0.6187}} \\
		&WaterNet \cite{liUnderwaterImageEnhancement2019} & 2.7543   & 0.5739   & 2.1274  & 0.5452   \\
		&UWCNN-typeI \cite{liUnderwaterScenePrior2020} & 2.3394  & 0.4862   & 1.5029   & 0.4454   \\
		&Ucolor \cite{liUnderwaterImageEnhancement2021}   & 2.5266   & 0.5407    & 1.85     & 0.5176  \\
		&U-Trans \cite{pengUshapeTransformerUnderwater2021} & 2.5246   & 0.5343  & 1.9507   & 0.5237   \\
		\multirow{-8}{*}{Whole \textbf{LSUI}} &Ours  &  2.6776  & 0.5874   & {\color[HTML]{C00000} \textbf{2.3681}} & 0.584   \\ \bottomrule                              
	\end{tabular}
\end{table}

Further, we conduct a no-reference evaluation comparison on \textbf{Test-C60} and \textbf{Test-S16}. The total statistical results are summarized in \autoref{tab:1}. As reported in \cite{bermanUnderwaterSingleImage2020}, UIQM and UCIQE are biased to serval features and insensitive to color artifacts and casts. Hence, they cannot determine everything while the visual comparison is necessary.

It should be noted that HL and IBLA produced obvious color deviations shown in \autoref{fig:6}, causing obvious failure. However, they reached the highest scores respectively in UIQM and UCIQE since deep learning-based UIE methods perform more weakly compared to the other two UIEs. Without these two methods, the proposed URSCT-UIE achieves the SOTA on UCIQE and is slightly inferior to WaterNet on UIQM.

Similarly,  the qualitative comparison is shown in \hyperref[sec:App]{\textbf{Appendix}}.
\section{Ablation Study}
\label{sec:AbS}
\begin{figure}[h]
	\centering
	\begin{minipage}[t]{0.48\textwidth}
		\centering
		\subfloat[\label{fig:a}]{
			\includegraphics[scale=0.2]{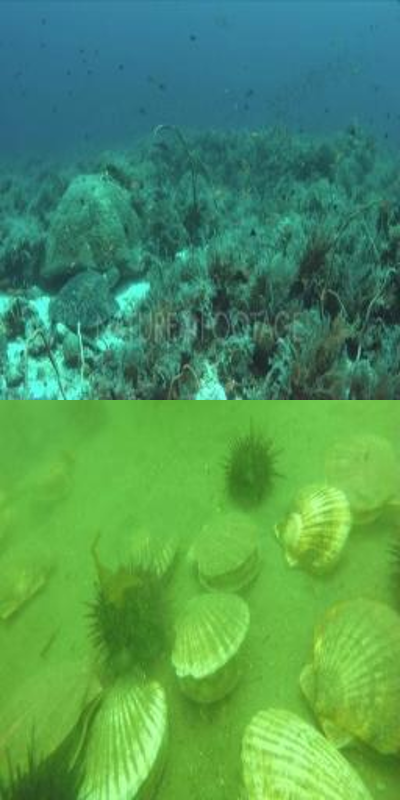}}
		\subfloat[\label{fig:b}]{
			\includegraphics[scale=0.2]{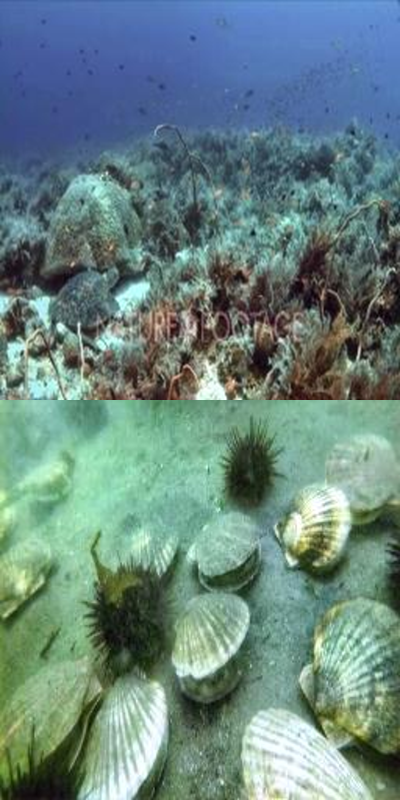}}
		\subfloat[\label{fig:c}]{
			\includegraphics[scale=0.2]{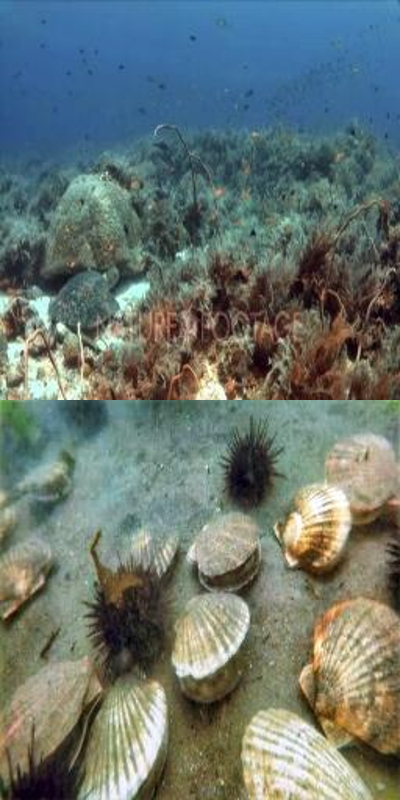}}
		\subfloat[\label{fig:d}]{
			\includegraphics[scale=0.2]{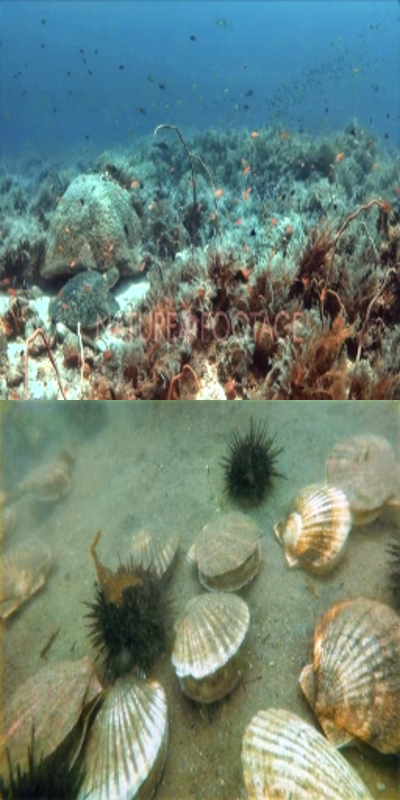}}
		\caption{Ablation study of the contributions of reinforced convolutions. (a) RAW. (b) The result of the origin. (c) The result of the Conv-typeI (proposed). (d) The result of the Conv-typeII.}
		\label{fig:10}
	\end{minipage}
	\quad
	\begin{minipage}[t]{0.48\textwidth}
		\centering
		\subfloat[\label{fig:a}]{
			\includegraphics[scale=0.2]{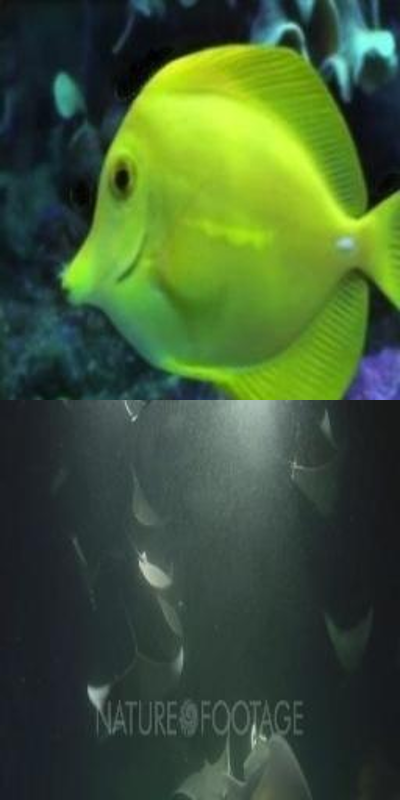}}
		\subfloat[\label{fig:b}]{
			\includegraphics[scale=0.2]{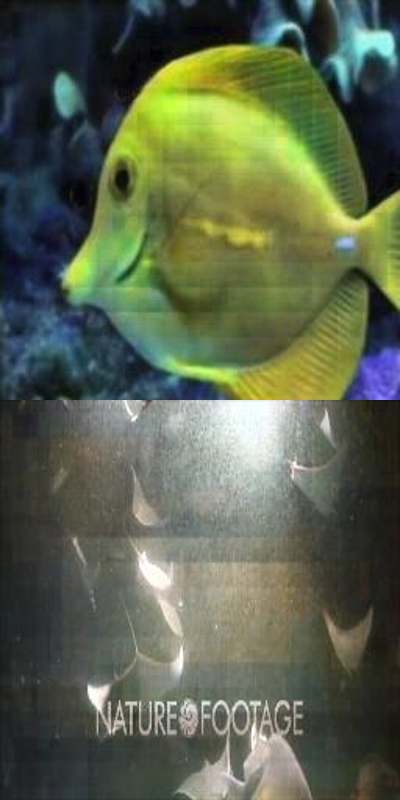}}
		\subfloat[\label{fig:c}]{
			\includegraphics[scale=0.2]{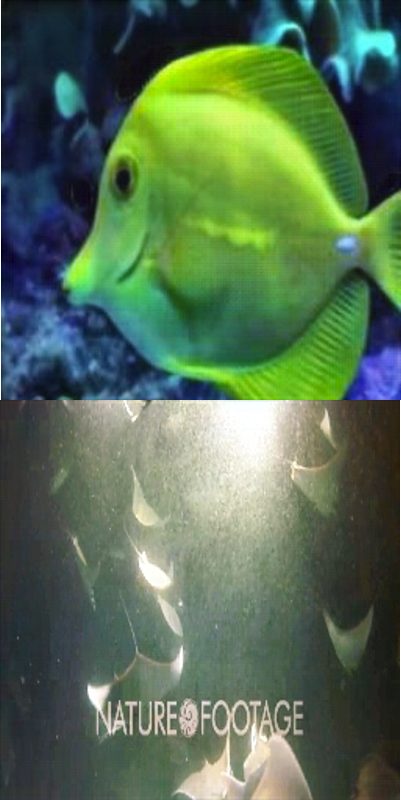}}
		\subfloat[\label{fig:d}]{
			\includegraphics[scale=0.2]{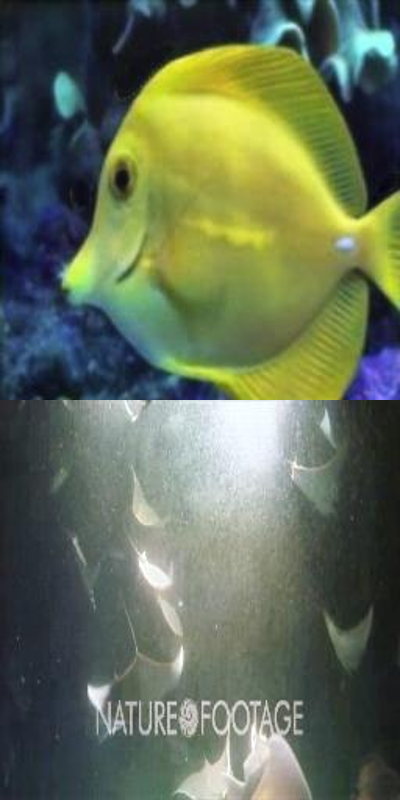}}
	\caption{Ablation study of the contributions of the proposed loss functions. (a) Raw. (b) $\mathcal{L} _{\mathrm{C}}$.(c) $\,\,\mathcal{L} _{\mathrm{C}}, \mathcal{L} _{\mathrm{M}}$. (d)$\,\,\mathcal{L} _{\mathrm{C}}, \mathcal{L} _{\mathrm{gd}}, \mathcal{L} _{\mathrm{M}}$ (proposed). }
	\label{fig:11}
	\end{minipage}
\end{figure}

The improvement of URSCT benefits greatly from the reinforced convolutions and redesigned loss function. Therefore, we conduct ablation studies with the model training on \textbf{Train-U}, which testing in \textbf{Test-L}. The experimental setting keeps unchanged with the experiments above. From \autoref{tab:5}, \autoref{fig:10}, and \autoref{fig:11}, one could see that: 1) Compared with the Conv-typeI, the lack of local attention in spatial causes the overexploitation of the overall blue channel information. The parallel mode of Conv-typeII wastes the information reinforced by the previous linear projection, while the series mode of Conv-typeI does. It shows that the proposed RSCTB is adept at escaping from blue and green artifacts, which is a common problem in UIE.
2) It is evident that the absence of gradient loss causes the network to have difficulty capturing gradient changes in color, which introduces the severe incoherency between small blocks and color deviation.

\begin{table}
	\caption{The statistical results of ablation study about modules and loss function on \textbf{Test-L}. The highest scores are marked in red.}
	\label{tab:5}
	\centering
	\begin{tabular}{ccccccc}
		\toprule
		&\multicolumn{3}{c}{\textbf{Module}}&\multicolumn{3}{c}{\textbf{Loss Function}} \\ \cmidrule{2-7}
		&Origin & Conv-typeI   & Conv-typeII & $\mathcal{L}_{\mathrm{C}} $ & $\mathcal{L}   _{\mathrm{C}}, \mathcal{L} _{\mathrm{M}}$ & $\mathcal{L}   _{\mathrm{C}}, \mathcal{L} _{\mathrm{gd}}, \mathcal{L} _{\mathrm{M}}$ \\ \midrule
		PSNR & 20.90  & {\color[HTML]{C00000} \textbf{22.32}} & 21.46    & 21.35  & 21.74    & {\color[HTML]{C00000}\textbf{22.32}}	\\ \midrule
		SSIM & 0.857  & {\color[HTML]{C00000} \textbf{0.871}} & 0.863   & 0.85   & 0.857    & {\color[HTML]{C00000} \textbf{0.862}} \\ \bottomrule
	\end{tabular}
\end{table}

%
%
%
%
%
%
\section{Conclusion}
In this paper, we presented a U-Net based Reinforced Swin-Convs Transformer for Underwater Image Enhancement method (URSCT-UIE). The proposed URSCT-UIE embedded the Swin Transformer into U-Net for improving the ability to capture the global dependency and redesigned the Swin Transformer, whose linear layer is replaced by convolutions to integrate the merit of CNN and keep the robustness in small datasets. Besides, combined with the redesigned multi-objective loss function, restoring performance and nuanced texture are further improved. The superior ability of the proposed URSCT-UIE to remove color artifacts and casts has been shown in the experiments on mainstream datasets. Furthermore, the ablation studies verified that the key improvement to the STB and the elaborately designed loss function is effective to face challenging underwater images. However, the proposed URSCT-UIE also met the failure in several special environments. In the future, we will attempt to combine some heads in the Transformer block with the pre-rea-channel-compensation mechanism and introduce convolutions as another branch paralleled with a better manner.

\newpage
\section*{References}
\begingroup
\renewcommand{\section}[2]{}%
\bibliographystyle{unsrt}
\bibliography{ref.bib}

\begin{thebibliography}{10}

\bibitem{zhaoConvolutionalNeuralNetwork2021}
Wanting Zhao, Hong Qi, Yu~Jiang, Chong Wang, and Fenglin Wei.
\newblock A convolutional neural network accelerator for real-time underwater
  image recognition of autonomous underwater vehicle.
\newblock {\em Proceedings of the Institution of Mechanical Engineers, Part I:
  Journal of Systems and Control Engineering}, 235(10):1839--1848, 2021.

\bibitem{congUnderwaterRobotSensing2021}
Yang Cong, Changjun Gu, Tao Zhang, and Yajun Gao.
\newblock Underwater robot sensing technology: {{A}} survey.
\newblock {\em Fundamental Research}, 1(3):337--345, 2021.

\bibitem{jianUnderwaterImageProcessing2021}
Muwei Jian, Xiangyu Liu, Hanjiang Luo, Xiangwei Lu, Hui Yu, and Junyu Dong.
\newblock Underwater image processing and analysis: {{A}} review.
\newblock {\em Signal Processing: Image Communication}, 91:116088, 2021.

\bibitem{baukKeyFeaturesAutonomous2021}
Sanja Bauk, Nexhat Kapidani, Jean-Philippe Boisgard, and {\v Z}arko Luk{\v
  s}i{\'c}.
\newblock Key features of the autonomous underwater vehicles for marine
  surveillance missions.
\newblock The 1st {{International Conference}} on {{Maritime Education}} and
  {{Development}}, pages 69--82, 2021.

\bibitem{liUnderwaterBiologicalDetection2022}
Aolun Li, Long Yu, and Shengwei Tian.
\newblock Underwater biological detection based on {{YOLOv4}} combined with
  channel attention.
\newblock {\em Journal of Marine Science and Engineering}, 10(4):469, 2022.

\bibitem{chenUnderwaterImageEnhancement2021}
Xuelei Chen, Pin Zhang, Lingwei Quan, Chao Yi, and Cunyue Lu.
\newblock Underwater image enhancement based on deep learning and image
  formation model.
\newblock {\em arXiv preprint arXiv:2101.00991}, 2021.

\bibitem{pengGeneralizationDarkChannel2018}
Yan-Tsung Peng, Keming Cao, and Pamela~C Cosman.
\newblock Generalization of the dark channel prior for single image
  restoration.
\newblock {\em IEEE Transactions on Image Processing}, 27(6):2856--2868, 2018.

\bibitem{wangNaturalnessPreservedEnhancement2013}
Shuhang Wang, Jin Zheng, Hai-Miao Hu, and Bo~Li.
\newblock Naturalness preserved enhancement algorithm for non-uniform
  illumination images.
\newblock {\em IEEE transactions on image processing}, 22(9):3538--3548, 2013.

\bibitem{bermanUnderwaterSingleImage2020}
Dana Berman, Deborah Levy, Shai Avidan, and Tali Treibitz.
\newblock Underwater single image color restoration using haze-lines and a new
  quantitative dataset.
\newblock {\em IEEE transactions on pattern analysis and machine intelligence},
  43(8):2822--2837, 2020.

\bibitem{fuTwostepApproachSingle2017}
Xueyang Fu, Zhiwen Fan, Mei Ling, Yue Huang, and Xinghao Ding.
\newblock Two-step approach for single underwater image enhancement.
\newblock 2017 International Symposium on Intelligent Signal Processing and
  Communication Systems ({{ISPACS}}), pages 789--794, 2017.

\bibitem{liSingleUnderwaterImage2016}
Chongyi Li, Jichang Quo, Yanwei Pang, Shanji Chen, and Jian Wang.
\newblock Single underwater image restoration by blue-green channels dehazing
  and red channel correction.
\newblock 2016 {{IEEE International Conference}} on {{Acoustics}}, {{Speech}}
  and {{Signal Processing}} ({{ICASSP}}), pages 1731--1735, 2016.

\bibitem{ancutiEnhancingUnderwaterImages2012}
Cosmin Ancuti, Codruta~Orniana Ancuti, Tom Haber, and Philippe Bekaert.
\newblock Enhancing underwater images and videos by fusion.
\newblock 2012 {{IEEE}} Conference on Computer Vision and Pattern Recognition,
  pages 81--88, 2012.

\bibitem{iqbalEnhancingLowQuality2010}
Kashif Iqbal, Michael Odetayo, Anne James, Rosalina~Abdul Salam, and Abdullah
  Zawawi~Hj Talib.
\newblock Enhancing the low quality images using unsupervised colour correction
  method.
\newblock 2010 {{IEEE International Conference}} on {{Systems}}, {{Man}} and
  {{Cybernetics}}, pages 1703--1709, 2010.

\bibitem{ghaniUnderwaterImageQuality2015}
Ahmad Shahrizan~Abdul Ghani and Nor Ashidi~Mat Isa.
\newblock Underwater image quality enhancement through integrated color model
  with {{Rayleigh}} distribution.
\newblock {\em Applied soft computing}, 27:219--230, 2015.

\bibitem{ancutiColorBalanceFusion2017}
Codruta~O Ancuti, Cosmin Ancuti, Christophe De~Vleeschouwer, and Philippe
  Bekaert.
\newblock Color balance and fusion for underwater image enhancement.
\newblock {\em IEEE Transactions on image processing}, 27(1):379--393, 2017.

\bibitem{ancutiColorChannelCompensation2019}
Codruta~O Ancuti, Cosmin Ancuti, Christophe De~Vleeschouwer, and Mateu Sbert.
\newblock Color channel compensation ({{3C}}): {{A}} fundamental pre-processing
  step for image enhancement.
\newblock {\em IEEE Transactions on Image Processing}, 29:2653--2665, 2019.

\bibitem{liUnderwaterImageEnhancement2019}
Chongyi Li, Chunle Guo, Wenqi Ren, Runmin Cong, Junhui Hou, Sam Kwong, and
  Dacheng Tao.
\newblock An underwater image enhancement benchmark dataset and beyond.
\newblock {\em IEEE Transactions on Image Processing}, 29:4376--4389, 2019.

\bibitem{liUnderwaterScenePrior2020}
Chongyi Li, Saeed Anwar, and Fatih Porikli.
\newblock Underwater scene prior inspired deep underwater image and video
  enhancement.
\newblock {\em Pattern Recognition}, 98:107038, 2020.

\bibitem{liUnderwaterImageEnhancement2021}
Chongyi Li, Saeed Anwar, Junhui Hou, Runmin Cong, Chunle Guo, and Wenqi Ren.
\newblock Underwater image enhancement via medium transmission-guided
  multi-color space embedding.
\newblock {\em IEEE Transactions on Image Processing}, 30:4985--5000, 2021.

\bibitem{pengUshapeTransformerUnderwater2021}
Lintao Peng, Chunli Zhu, and Liheng Bian.
\newblock U-shape transformer for underwater image enhancement.
\newblock {\em arXiv preprint arXiv:2111.11843}, 2021.

\bibitem{dudhaneEndtoendNetworkImage2020}
Akshay Dudhane, Prashant~W Patil, and Subrahmanyam Murala.
\newblock An end-to-end network for image de-hazing and beyond.
\newblock {\em IEEE Transactions on Emerging Topics in Computational
  Intelligence}, 2020.

\bibitem{akkaynakWhatSpaceAttenuation2017}
Derya Akkaynak, Tali Treibitz, Tom Shlesinger, Yossi Loya, Raz Tamir, and David
  Iluz.
\newblock What is the space of attenuation coefficients in underwater computer
  vision?
\newblock Proceedings of the {{IEEE Conference}} on {{Computer Vision}} and
  {{Pattern Recognition}}, pages 4931--4940, 2017.

\bibitem{wangDeepCNNMethod2017}
Yang Wang, Jing Zhang, Yang Cao, and Zengfu Wang.
\newblock A deep {{CNN}} method for underwater image enhancement.
\newblock 2017 {{IEEE International Conference}} on {{Image Processing}}
  ({{ICIP}}), pages 1382--1386, 2017.

\bibitem{wangUIEC2NetCNNbased2021}
Yudong Wang, Jichang Guo, Huan Gao, and Huihui Yue.
\newblock {{UIEC}}\textbackslash\textasciicircum\{\} 2-{{Net}}: {{CNN-based}}
  underwater image enhancement using two color space.
\newblock {\em Signal Processing: Image Communication}, 96:116250, 2021.

\bibitem{lyuEfficientLearningbasedMethod2022}
Zhangkai Lyu, Andrew Peng, Qingwei Wang, and Dandan Ding.
\newblock An efficient learning-based method for underwater image enhancement.
\newblock {\em Displays}, page 102174, 2022.

\bibitem{tanEfficientnetv2SmallerModels2021}
Mingxing Tan and Quoc Le.
\newblock Efficientnetv2: {{Smaller}} models and faster training.
\newblock International {{Conference}} on {{Machine Learning}}, pages
  10096--10106, 2021.

\bibitem{vaswaniAttentionAllYou2017}
Ashish Vaswani, Noam Shazeer, Niki Parmar, Jakob Uszkoreit, Llion Jones,
  Aidan~N Gomez, {\L}ukasz Kaiser, and Illia Polosukhin.
\newblock Attention is all you need.
\newblock {\em Advances in neural information processing systems}, 30, 2017.

\bibitem{dosovitskiyImageWorth16x162020}
Alexey Dosovitskiy, Lucas Beyer, Alexander Kolesnikov, Dirk Weissenborn,
  Xiaohua Zhai, Thomas Unterthiner, Mostafa Dehghani, Matthias Minderer, Georg
  Heigold, Sylvain Gelly, et~al.
\newblock An image is worth 16x16 words: {{Transformers}} for image recognition
  at scale.
\newblock {\em arXiv preprint arXiv:2010.11929}, 2020.

\bibitem{liuSwinTransformerHierarchical2021}
Ze~Liu, Yutong Lin, Yue Cao, Han Hu, Yixuan Wei, Zheng Zhang, Stephen Lin, and
  Baining Guo.
\newblock Swin transformer: {{Hierarchical}} vision transformer using shifted
  windows.
\newblock Proceedings of the {{IEEE}}/{{CVF International Conference}} on
  {{Computer Vision}}, pages 10012--10022, 2021.

\bibitem{ronnebergerUnetConvolutionalNetworks2015}
Olaf Ronneberger, Philipp Fischer, and Thomas Brox.
\newblock U-net: {{Convolutional}} networks for biomedical image segmentation.
\newblock International {{Conference}} on {{Medical}} Image Computing and
  Computer-Assisted Intervention, pages 234--241, 2015.

\bibitem{shangPerceptualExtremeSuperresolution2020}
Taizhang Shang, Qiuju Dai, Shengchen Zhu, Tong Yang, and Yandong Guo.
\newblock Perceptual extreme super-resolution network with receptive field
  block.
\newblock Proceedings of the {{IEEE}}/{{CVF Conference}} on {{Computer Vision}}
  and {{Pattern Recognition Workshops}}, pages 440--441, 2020.

\bibitem{raffelExploringLimitsTransfer2019}
Colin Raffel, Noam Shazeer, Adam Roberts, Katherine Lee, Sharan Narang, Michael
  Matena, Yanqi Zhou, Wei Li, and Peter~J Liu.
\newblock Exploring the limits of transfer learning with a unified text-to-text
  transformer.
\newblock {\em arXiv preprint arXiv:1910.10683}, 2019.

\bibitem{isolaImagetoimageTranslationConditional2017}
Phillip Isola, Jun-Yan Zhu, Tinghui Zhou, and Alexei~A Efros.
\newblock Image-to-image translation with conditional adversarial networks.
\newblock Proceedings of the {{IEEE}} Conference on Computer Vision and Pattern
  Recognition, pages 1125--1134, 2017.

\bibitem{bruhnLucasKanadeMeets2005}
Andr{\'e}s Bruhn, Joachim Weickert, and Christoph Schn{\"o}rr.
\newblock Lucas/{{Kanade}} meets {{Horn}}/{{Schunck}}: {{Combining}} local and
  global optic flow methods.
\newblock {\em International journal of computer vision}, 61(3):211--231, 2005.

\bibitem{mathieuDeepMultiscaleVideo2015}
Michael Mathieu, Camille Couprie, and Yann LeCun.
\newblock Deep multi-scale video prediction beyond mean square error.
\newblock {\em arXiv preprint arXiv:1511.05440}, 2015.

\bibitem{wangMultiscaleStructuralSimilarity2003}
Zhou Wang, Eero~P Simoncelli, and Alan~C Bovik.
\newblock Multiscale structural similarity for image quality assessment.
\newblock volume~2 of {\em The {{Thrity-Seventh Asilomar Conference}} on
  {{Signals}}, {{Systems}} \textbackslash\& {{Computers}}, 2003}, pages
  1398--1402, 2003.

\bibitem{wangImageQualityAssessment2004}
Zhou Wang, Alan~C Bovik, Hamid~R Sheikh, and Eero~P Simoncelli.
\newblock Image quality assessment: From error visibility to structural
  similarity.
\newblock {\em IEEE transactions on image processing}, 13(4):600--612, 2004.

\bibitem{loshchilovDecoupledWeightDecay2017}
Ilya Loshchilov and Frank Hutter.
\newblock Decoupled weight decay regularization.
\newblock {\em arXiv preprint arXiv:1711.05101}, 2017.

\bibitem{islamSimultaneousEnhancementSuperresolution2020}
Md~Jahidul Islam, Peigen Luo, and Junaed Sattar.
\newblock Simultaneous enhancement and super-resolution of underwater imagery
  for improved visual perception.
\newblock {\em arXiv preprint arXiv:2002.01155}, 2020.

\bibitem{korhonenPeakSignaltonoiseRatio2012}
Jari Korhonen and Junyong You.
\newblock Peak signal-to-noise ratio revisited: {{Is}} simple beautiful?
\newblock 2012 {{Fourth International Workshop}} on {{Quality}} of {{Multimedia
  Experience}}, pages 37--38, 2012.

\bibitem{yangUnderwaterColorImage2015}
Miao Yang and Arcot Sowmya.
\newblock An underwater color image quality evaluation metric.
\newblock {\em IEEE Transactions on Image Processing}, 24(12):6062--6071, 2015.

\bibitem{panettaHumanvisualsysteminspiredUnderwaterImage2015}
Karen Panetta, Chen Gao, and Sos Agaian.
\newblock Human-visual-system-inspired underwater image quality measures.
\newblock {\em IEEE Journal of Oceanic Engineering}, 41(3):541--551, 2015.

\bibitem{chenMFFNUnderwaterSensing2021}
Renzhang Chen, Zhanchuan Cai, and Wei Cao.
\newblock {{MFFN}}: {{An}} underwater sensing scene image enhancement method
  based on multiscale feature fusion network.
\newblock {\em IEEE Transactions on Geoscience and Remote Sensing}, 60:1--12,
  2021.

\end{thebibliography}
\endgroup

\section*{Checklist}


\begin{enumerate}

\item For all authors...
\begin{enumerate}
  \item Do the main claims made in the abstract and introduction accurately reflect the paper's contributions and scope?
    \answerYes{}
  \item Did you describe the limitations of your work?
    \answerYes{}
  \item Did you discuss any potential negative societal impacts of your work?
    \answerYes{}
  \item Have you read the ethics review guidelines and ensured that your paper conforms to them?
    \answerYes{}
\end{enumerate}

\item If you are including theoretical results...
\begin{enumerate}
  \item Did you state the full set of assumptions of all theoretical results?
    \answerNA{}
        \item Did you include complete proofs of all theoretical results?
    \answerNA{}
\end{enumerate}

\item If you ran experiments...
\begin{enumerate}
  \item Did you include the code, data, and instructions needed to reproduce the main experimental results (either in the supplemental material or as a URL)?
    \answerYes{}
  \item Did you specify all the training details (e.g., data splits, hyperparameters, how they were chosen)?
    \answerYes{}
        \item Did you report error bars (e.g., with respect to the random seed after running experiments multiple times)?
    \answerYes{}
        \item Did you include the total amount of compute and the type of resources used (e.g., type of GPUs, internal cluster, or cloud provider)?
    \answerYes{}
\end{enumerate}

\item If you are using existing assets (e.g., code, data, models) or curating/releasing new assets...
\begin{enumerate}
  \item If your work uses existing assets, did you cite the creators?
    \answerYes{}
  \item Did you mention the license of the assets?
    \answerNA{}
  \item Did you include any new assets either in the supplemental material or as a URL?
    \answerNA{}
  \item Did you discuss whether and how consent was obtained from people whose data you're using/curating?
    \answerNA{}
  \item Did you discuss whether the data you are using/curating contains personally identifiable information or offensive content?
    \answerNA{}
\end{enumerate}

\item If you used crowdsourcing or conducted research with human subjects...
\begin{enumerate}
  \item Did you include the full text of instructions given to participants and screenshots, if applicable?
    \answerNA{}
  \item Did you describe any potential participant risks, with links to Institutional Review Board (IRB) approvals, if applicable?
    \answerNA{}
  \item Did you include the estimated hourly wage paid to participants and the total amount spent on participant compensation?
    \answerNA{}
\end{enumerate}

\end{enumerate}

%
%

\section*{Appendix}
\label{sec:App}
We conduct evaluation on the cross-dataset as the additional experiments. In training, we used four mainstream benchmarks: 800 paired real underwater images on the UIEB dataset \cite{liUnderwaterImageEnhancement2019}, denoted as \textbf{Train-U}; 4500 paired real underwater images randomly divided from the LSUI dataset \cite{pengUshapeTransformerUnderwater2021}, denoted as \textbf{Train-L}; 1500 paired real underwater images from the UFO-120 dataset \cite{islamSimultaneousEnhancementSuperresolution2020}, denoted as \textbf{Train-UF}. In testing, apart from the rest of the corresponding dataset, each trained model will test the other three datasets for generalization capability. Specifically, \textbf{Test-U} denoted the whole UIEB dataset or the rest of 90 paired data from the UIEB dataset if the training dataset is \textbf{Train-U}; \textbf{Test-L} denoted the whole LSUI dataset or the rest 504 paired data from dataset LSUI if the training dataset is \textbf{Train-U}; \textbf{Test-UF} denoted the whole UFO-120 dataset or the preassigned 120 paired data from the UFO-120 dataset if the training dataset is \textbf{Train-UF}.

The the quantitative comparison of three cross-datasets are summarized in \autoref{tab:4}.

\begin{table}[h]
	\caption{The three cross-datasets evaluations in terms of average PSNR and SSIM Values.}
	\label{tab:4}
	\centering
	\begin{tabular}{ccccccc}
		\toprule
		& \multicolumn{2}{c}{\textbf{Test-U}}                   & \multicolumn{2}{c}{\textbf{Test-L}}                   & \multicolumn{2}{c}{\textbf{Test-UF}} \\ \cmidrule{2-7}
		\multirow{-2}{*}{\textbf{Training}} & PSNR  & SSIM                                 & PSNR                                  & SSIM & PSNR          & SSIM        \\ \midrule
		\textbf{Train-U}                         & 22.72 & 0.91 & 22.32 & 0.87 & 19.09         & 0.74        \\
		\textbf{Train-L}                         & 26.70 & 0.93 & 29.31 & 0.93 & 26.26         & 0.83        \\
		\textbf{Train-UF}                        & 15.30 & 0.70 & 21.92 & 0.81 & 27.03         & 0.84      \\ \bottomrule 
	\end{tabular}
\end{table}

\begin{figure}[h]
	\centering
	\subfloat[UIEB\label{fig:a}]{
		\includegraphics[scale=0.3]{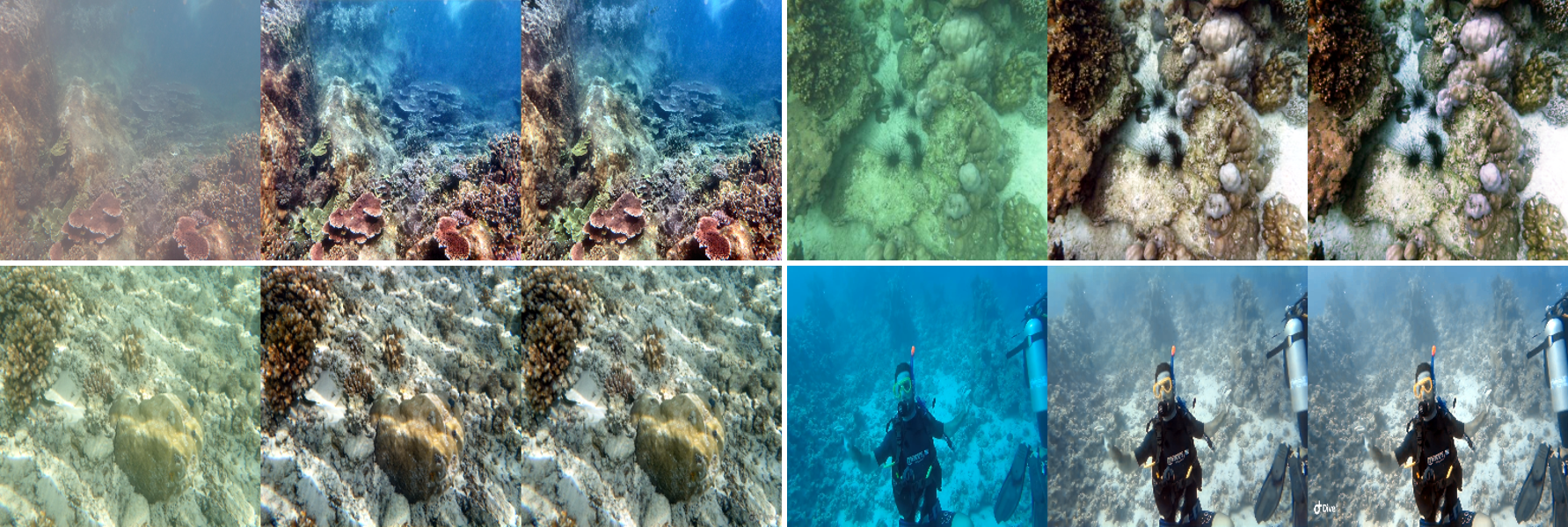}} \\
	\subfloat[LSUI\label{fig:c}]{
		\includegraphics[scale=0.3]{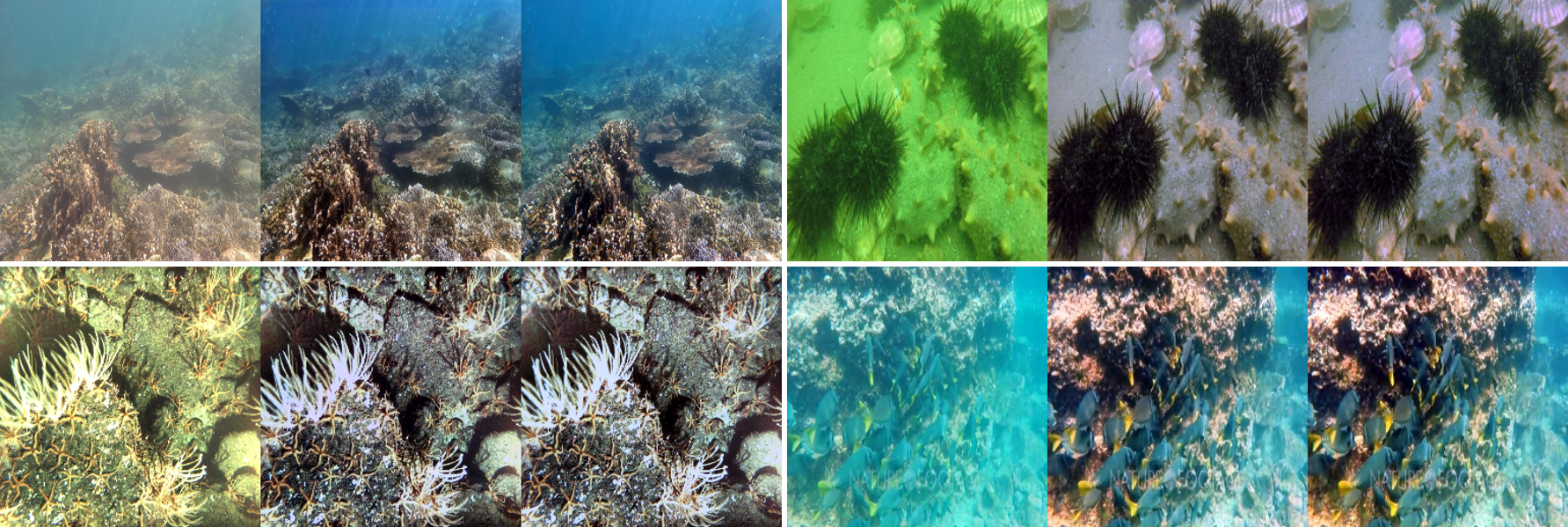}} \\
	\subfloat[UFO-120\label{fig:e}]{
		\includegraphics[scale=0.3]{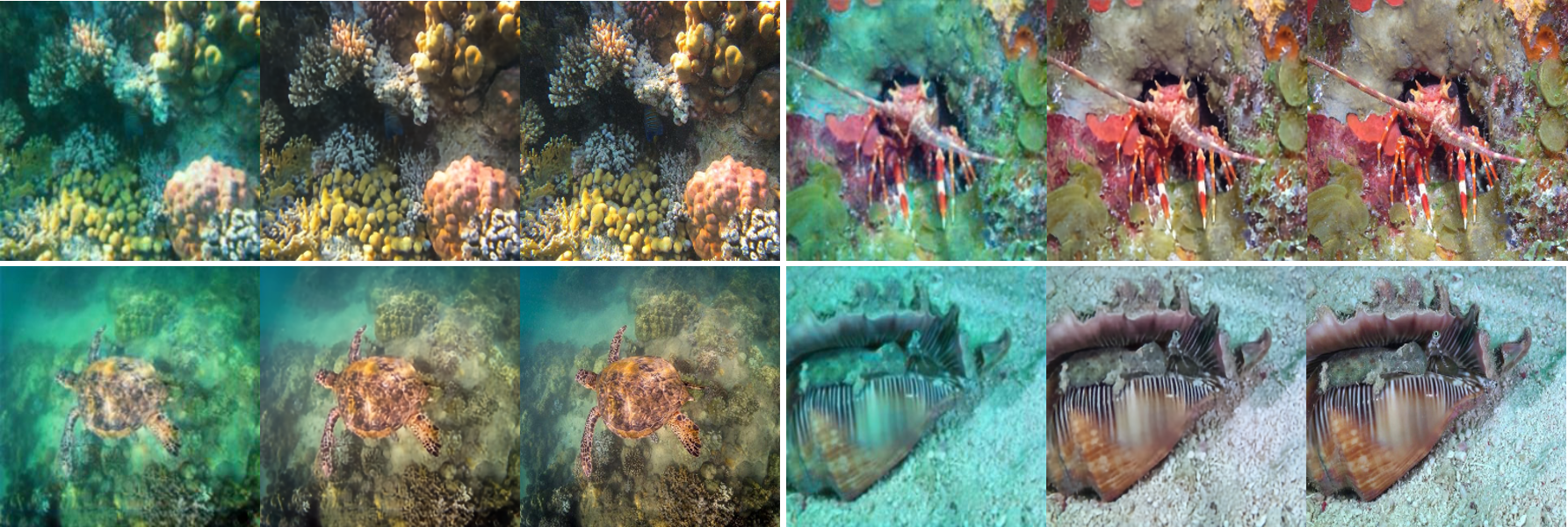}} \\
	
	\caption{The restored testing results in the corresponding training dataset. In each case, there is RAW, the result of our URSCT, and the ground truth from left to right.}
	\label{fig:8}
\end{figure}

The UIEB dataset and the LSUI dataset are more similar in style and even have some overlapping data, giving closer quantitative results. Four representative scenes, namely haze (top left), green (top right), yellowish (bottom left), and blue (bottom right) scenes, are selected. Apart from enhancement, the UFO-120 dataset is also used for super-resolution. Hence, we selected four scenes, i.e., high requirements for image texture details (top left), disturbing colors (top right), green cover (bottom left), and blue cover (bottom right), shown in \autoref{fig:8}.

\begin{figure}[h]
	\centering
	\subfloat[\textbf{Train-U}, \textbf{Test-L}\label{fig:a}]{
		\includegraphics[scale=0.3]{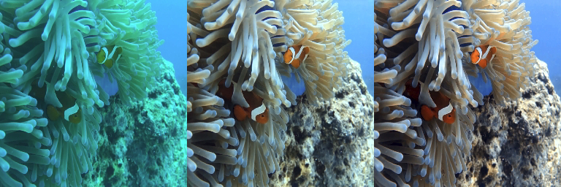}}
	\subfloat[\textbf{Train-U}, \textbf{Test-UF}\label{fig:a}]{
		\includegraphics[scale=0.3]{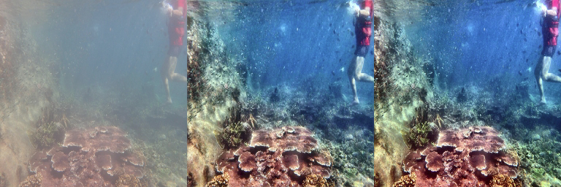}} \\
	\subfloat[\textbf{Train-L}, \textbf{Test-U}\label{fig:a}]{
		\includegraphics[scale=0.3]{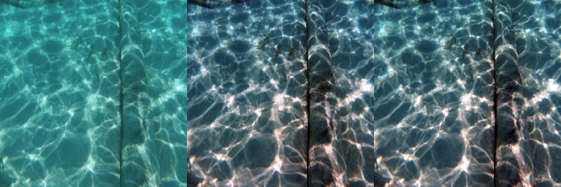}}
	\subfloat[\textbf{Train-L}, \textbf{Test-UF}\label{fig:a}]{
		\includegraphics[scale=0.3]{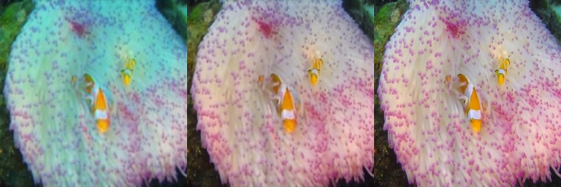}} \\
	\subfloat[\textbf{Train-UF}, \textbf{Test-U}\label{fig:a}]{
		\includegraphics[scale=0.3]{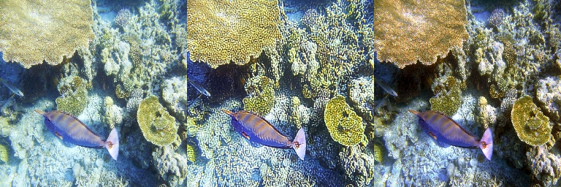}}
	\subfloat[\textbf{Train-UF}, \textbf{Test-L}\label{fig:a}]{
		\includegraphics[scale=0.3]{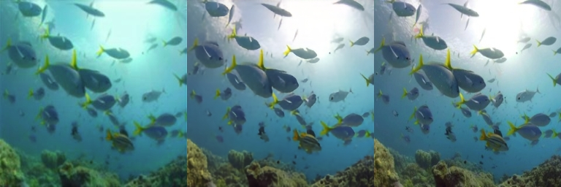}} \\
	
	\caption{The restored testing results in the cross-dataset. In each case, there is RAW, the result of our URSCT, and the ground truth from left to right.}
	\label{fig:9}
\end{figure}

Finally, the representative enhanced results on the cross-dataset are shown in \autoref{fig:9}, whose training datasets are different from testing datasets.

\end{document}